\documentclass[journal]{IEEEtran}
\usepackage{setspace}
\usepackage{float}
\usepackage[fleqn]{amsmath}
\usepackage{enumerate}%\begin{enumerate}[i] % reacts to 1, a, A, i, and I
\usepackage{mathrsfs}
\usepackage{makeidx}         % allows index generation
\usepackage{graphicx}        % standard LaTeX graphics tool
\usepackage{multicol}        % used for the two-column index
\usepackage{subfigure}
\usepackage{epsfig,amssymb,latexsym}
\usepackage{psfrag}
\usepackage{algorithm}
\usepackage{algorithmic}
\usepackage{url}

\newtheorem{assumption}{Assumption}%[section]
\newtheorem{remark}{Remark}%[section]
\newtheorem{theorem}{Theorem}%[section]
\newtheorem{lemma}{Lemma}%[section]

\newcommand{\be}{\begin{equation}}
\newcommand{\ee}{\end{equation}}
\newcommand{\bea}{\begin{eqnarray}}
\newcommand{\eea}{\end{eqnarray}}
\newcommand{\bal}{\begin{align}}
\newcommand{\eal}{\end{align}}
\newcommand{\ba}{\begin{array}}
\newcommand{\ea}{\end{array}}
\newcommand{\bc}{\begin{center}}
\newcommand{\ec}{\end{center}}

\allowdisplaybreaks
\ifCLASSINFOpdf
\else
\fi
\begin{document}
\title{Adaptive Control for Marine Vessels Against Harsh Environmental Variation}

\author{Fangwen~Tu,
        Shuzhi~Sam~Ge,~\IEEEmembership{Fellow,~IEEE,}
        Yoo~Sang~Choo,
        and~Chang~Chieh~Hang,~\IEEEmembership{Fellow,~IEEE}% <-this % stops a space
%\thanks{F. Tu is with the Department of Electrical and Computer Engineering, National
%University of Singapore, Singapore 117576 (e-mail: $\text{fangwen\_tu@hotmail.com}$).}% <-this % stops a space
%\thanks{S. S. Ge is with the Department of Electrical and Computer Engineering, National
%University of Singapore, Singapore 117576 and also with the Social Robotics Lab, Interactive Digital
%Media Institute (IDMI), National University of Singapore, Singapore 117576, (e-mail: samge@nus.edu.sg).}% <-this % stops a space
%\thanks{Y. S. Choo is with the Centre for Offshore Research and Engineering, National
%University of Singapore, 117576 Singapore (e-mail: cvecys@nus.edu.sg).}
%\thanks{C. C. Hang is with the Department of Electrical and Computer Engineering, National
%University of Singapore, Singapore 117576 (e-mail: etmhead@nus.edu.sg).}
}

\markboth{IEEE TRANSACTIONS ON SYSTEMS, MAN, AND CYBERNETICS: SYSTEMS,~Vol.~x, No.~x, x~2017}%
{Shell \MakeLowercase{\textit{et al.}}: Bare Demo of IEEEtran.cls for Journals}



\maketitle

\begin{abstract}
In this paper, robust control with sea state observer and dynamic thrust allocation is proposed for the Dynamic Positioning (DP) of an accommodation vessel in the presence of unknown hydrodynamic force variation and the input time delay. In order to overcome the huge force variation due to the adjoining Floating Production Storage and Offloading (FPSO) and accommodation vessel, a novel sea state observer is designed. The sea observer can effectively monitor the variation of the drift wave-induced force on the vessel and activate Neural Network (NN) compensator in the controller when large wave force is identified. Moreover, the wind drag coefficients can be adaptively approximated in the sea observer so that a feedforward control can be achieved. Based on this, a robust constrained control is developed to guarantee a safe operation. The time delay inside the control input is also considered. Dynamic thrust allocation module is presented to distribute the generalized control input among azimuth thrusters. Under the proposed sea observer and control, the boundedness of all the closed-loop signals are demonstrated via rigorous Lyapunov analysis. A set of simulation studies are conducted to verify the effectiveness of the proposed control scheme.\\
\end{abstract}


% \fi
\IEEEpeerreviewmaketitle

\section{Introduction}
\begin{figure}[htb]
  \centering
  \includegraphics[width=65mm,height=55mm]{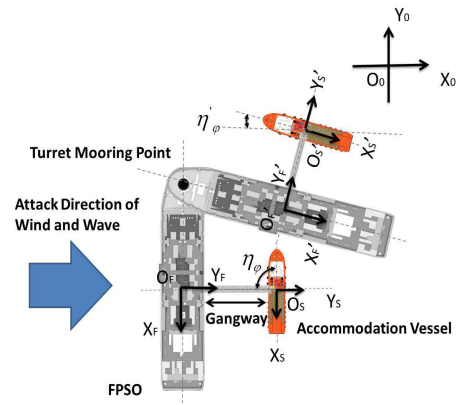}
  \caption{Definition of the coordinate system} \label{structure}
\end{figure}

\begin{figure*}[htb]
  \centering
  \includegraphics[width=140mm,height=55mm]{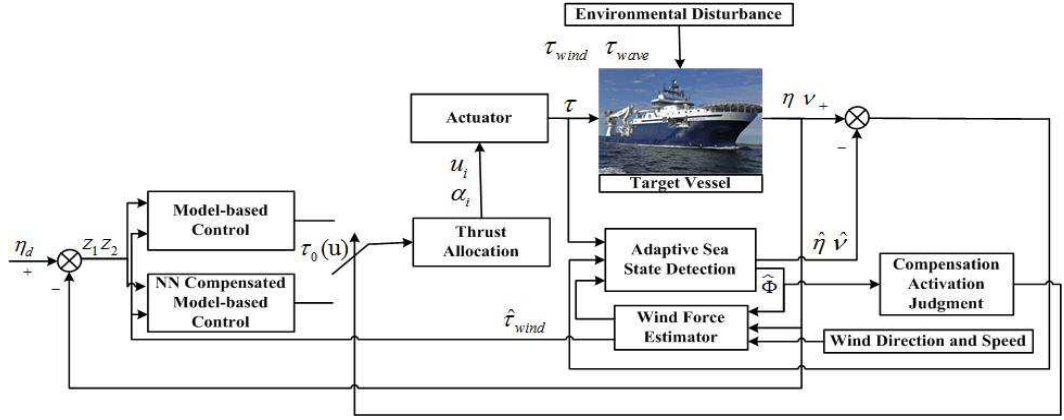}
  \caption{Diagram of sea state observer, controller and thrust allocator} \label{sysdiag}
\end{figure*}

FPSOs unit are highly demanded to produce, process hydrocarbons and store oil in marine industry. At the same time, Accommodation Vessels (AV) which can provide the space for logistic support and open deck space in deep sea environment is needed to handle the maintenance related work offshore. In this way, these AVs must ensure connected for continuous personnel and equipments transfer through gangway. Thus, the motivation of this paper is to design a DP system to allow the AVs to maintain proper relative position and heading under varying environmental situations.\\
% \begin{figure}[htb]
%  % Requires \usepackage{graphicx}
%  \centering
%  \includegraphics[width=90mm,height=55mm]{accvel.eps}
%  \caption{Structure of typical accommodation vessel} \label{accvel}
%\end{figure}
\indent One of the most significant phenomena during the operation is the hydrodynamic interaction between the two vessels. This strong influence is called shielding effect \cite{rampazzo2011numerical} which results in huge environmental force variation. The ocean waves can propagate in multiple directions. Once the smaller accommodation vessel situates in the downstream shadow of FPSO as shown in Fig. \ref{structure}, the large FPSO would protect the smaller vessel in the vicinity. Consequently, the vessel only receives small wave-induced force. When the vessel moves out of the shadow, the environmental loads on the AV would increase. Thus, it is a very challenging to keep a fixed relative position and heading under this variation. In order to alarm the it, for the first time, a novel sea state observer is proposed. The observer is motivated by the fault diagnosis process in fault tolerant control \cite{van2015robust} \cite{van2013robust}. Different from traditional fault observer, the sea state observer is able to adaptively estimate the wind force and moment. The estimated force and moment is used for a feedforward control to counteract the wind effect on the vessel. Based on this, the detection of the shielding effect is not only judged by the residual between actual system states and estimated states, but, the estimated wind drag coefficients are selected as the indicator of the shielding effect due to the over-estimation phenomena. After huge wave-induced force and moment are detected, NNs are applied in both sea observer and controller to compensate the wave force.\\
\indent Additionally, to ensure the extended length of gangway between an AV and the FPSO not exceed the limit stroke, the tracking errors must be regulated. In \cite{tee2011control}, Barrier Lyapunov Function (BLF) method was proposed to handle output constraint. Compared to other schemes, BLF needs less restrictive initial conditions and does not require the explicit system solution. A general framework to handle the prescribed performance tracking problem for strict feedback systems were proposed in \cite{bechlioulis2011robust}. Apart from tracking error constraint, the input delay existing in the trusters can severely degrade the control performance. The delay is mainly caused by the long response time of the thruster driver \cite{zhao2014robust}. Thus, it is necessary to take the input delay into consideration for the control design. Much research has been done to cope with input delay for linear system \cite{bekiaris2010stabilization} \cite{jankovic2009forwarding}. However, the nonlinearity of the vessel systems bring more challenges to the control design. In \cite{zhu2010adaptive} \cite{zhu2012new}, an adaptive tracking control scheme has been developed for a class of multi-input and multi-output (MIMO) nonlinear system with input delay. A virtual observer is constructed as an auxiliary system to convert the input delay system into a non-delayed one. A robust saturation control approach for vibration suppression of building structures with input delay is presented in \cite{du2011actuator}. This control is able to handle bounded time-varying input delay. But integrating tracking error constraint with input delay is seldom studied, especially for nonlinear systems. Therefore, in this paper, in order to guarantee a smooth and safe operation, both of these two requirements need to be considered simultaneously.\\
\indent In this paper, we consider an AV with 6 azimuth thrusters which can produce forces in all directions. The aim of thrust allocation module is to distribute the desired control effort among the trusters, i.e., to solve the required rotation angle and output thrust for each thruster. The overactuated propelling system makes an optimization problem. In reality, dynamic allocation is needed since the formulation of the optimization problem depends on the earlier allocation results. Moreover, due to the deployment of azimuth thrusters, the optimization problem becomes a nonconvex one \cite{fossen2006survey}. Therefore, it is hard to utilize the traditional iterative numerical optimization method to search the solution. Since we always hope to search the optimal solution in the neiborhood of current thruster state (i.e. rotated angle and produced thrust), a method of local linearization \cite{johansen2004constrained} is proper and applicable to convert the nonconvex problem into a local convex one. Sequently, various methods such as linear programming \cite{dantzig1998linear} and NN dynamic solvers can be applied \cite{zhang2002dual}. Although thrust allocation problem have been extensively researched, few research results are available to combine thruster-thruster interaction and other thruster property constraints together. In this manner, a more intact dynamic characteristic of the thruster is considered. The block diagram of the overall DP system can be found in Fig. \ref{sysdiag}. \\
\indent The contributions of this paper is three-fold.
\begin{itemize}
\item[(i)] A novel model-based adaptive sea state observer is developed to alarm the huge environmental force variation and at the same time adaptively approximate the wind force and moment for feedforward compensation.
\item[(ii)] Robust adaptive control is proposed in combination with predictor-based method and symmetric BLF to handle constant control input delay and output tracking error constraints simultaneously. In addition, NN is employed for the compensation of force variation.
\item[(iii)] Both thruster-thruster interaction and other truster property are considered in the thrust allocation module. After locally convex reformulation, LVI-based Primal Dual Neural Network (LVI-PDNN) solver is designed to search the optimal solution accurately.
\end{itemize}
%The organization of this paper is as follows. Section \uppercase\expandafter{\romannumeral2} describes the problem formulation for DP control of FPSO-AV system. Section \uppercase\expandafter{\romannumeral3} presents the design of model-based sea state observer and the wind estimator. NN based robust adaptive tracking control considering input delay and error constraints are brought forward in Section \uppercase\expandafter{\romannumeral4}. Section \uppercase\expandafter{\romannumeral5} introduces the dynamic thrust allocation scheme for individual thruster. In Section \uppercase\expandafter{\romannumeral6}, extensive simulations of the proposed observer, controller and allocator are conducted to demonstrate the effectiveness of our approach. Section \uppercase\expandafter{\romannumeral7} gives some concluding remarks.
\section{Problem Formulation}
DP control is designed for FPSO-AV system operated under shielding effect as shown in Fig. \ref{structure}. The global frame ($X_0, O_0, Y_0$) is defined with the origin fixed at a certain point on sea level. The local frame of FPSO ($X_F, O_F, Y_F$) is a moving coordinate system with its origin fixed at the midship point in the water line. $X_F$ axis is the longitudinal axis which points to the stern of the ship. $Y_F$ is the transversal axis which directs to the starboard. The body frame of AV ($X_s, O_s. Y_s$) is defined very similar with that of the FPSO. Due to the turret mooring system and the exogenous environmental forces, the FPSO will make slow yaw motion about the turret pivot point. Thus, the AV is supposed to achieve corresponding plane motion and rotation to ensure a fixed relative position and orientation with FPSO. Let $\eta=[\eta_{\text{x}}, \eta_{\text{y}}, \eta_{\psi}]^T$ represents the earth-fixed position and heading of target vessel. The alongship, athwartship and rotational velocity are expressed by vector $\nu=[u_{\text{x}}, v_{\text{y}}, r_{\psi}]^T$. Referring to \cite{fossen2011handbook}, the low frequency (LF) dynamic model of the vessel is considered as follows.
\begin{eqnarray}
\label{overallmodel1}
\dot{\eta}=J(\eta_{\psi})\nu
\end{eqnarray}
\begin{equation}
\label{overallmodel2}
M\dot{\nu}+C(\nu)\nu+D(\nu)\nu+g(\eta)=\tau(t-t_d)+\gamma(t-T)\tau_{\text{wave}}+\tau_{\text{wind}}+d
\end{equation}
where $J(\eta_{\psi})$ is the rotation matrix defined as
\begin{eqnarray}
J(\eta_{\psi})=\begin{bmatrix}\text{cos}(\eta_{\psi})&\text{sin}(\eta_{\psi})& 0\\-\text{sin}(\eta_{\psi})&\text{cos}(\eta_{\psi})&0\\0&0&1\end{bmatrix}
\end{eqnarray}
$M=M_{\text{RB}}+M_{\text{A}} \in \mathbb{R}^{3 \times 3}$ is a known diagonal inertia matrix which is the sum of rigid body inertia and added mass. In DP control design, the inertia matrix is usually considered as a constant matrix \cite{fossen1994guidance} \cite{fossen2011handbook}. $C(\nu)=C_{\text{RB}}(\nu)+C_{\text{A}}(\nu)$ is the matrix of Coriolis and centripetal. $D(\nu)$ and $g(\eta)$ are the damping matrix and restoring force respectively. $d \in \mathbb{R}^{3}$ is the time-varying unknown external disturbance and unmodeled dynamics. $\tau(t-t_d)\in \mathbb{R}^{3}$ denotes the generalized control input with known constant time delay $t_d \in \mathbb{R}$. $\tau_{\text{wave}}\in \mathbb{R}^{3\times1}$ and $\tau_{\text{wind}}\in \mathbb{R}^{3\times1}$ represent the wave and wind force/moment. $\gamma(t-T)\tau_{\text{wave}}$ describes the hydrodynamic force variation with $T$ denoting the an uncertain moment that the vessel starts to be subjected to the wave force. The function $\gamma(t-T)$ is defined as
\begin{align}
\label{gammdef}
\gamma(t-T)=\begin{cases} \begin{array}{ll}0, \quad &\text{if} \quad t<T\\ \chi(t,T,t_{\text{T}}), \quad &\text{if} \quad t \geq T \end{array} \end{cases}
\end{align}
where
\begin{align}
\label{gammdef}
\chi(t,T,t_{\text{T}})=\begin{cases} \begin{array}{ll}\displaystyle\frac{T-t}{t_{\text{T}}}, \quad &\text{if} \quad t-T<t_{\text{T}}\\ 1, \quad &\text{if} \quad t-T\geq t_{\text{T}} \end{array} \end{cases}
\end{align}
$t_{\text{T}}$ represents the shielding time. The expression of wind force and moment in surge, sway and yaw are as follows \cite{fossen2011handbook}.
\begin{align}
\hspace{-5mm}
\notag\tau_{\text{wind}}=&\begin{bmatrix}0.5\rho_{\text{air}}C_{\text{x}}\cos(\eta_{\psi}-\beta_w)V^2_{\text{w}}A_{T}\\0.5\rho_{\text{air}}C_{\text{y}}\text{sin}(\eta_{\psi}-\beta_w)V^2_{\text{w}}A_{L}\\0.5\rho_{\text{air}}C_{\text{N}}\text{sin}[2(\eta_{\psi}-\beta_w)]V^2_{\text{w}}A_{L}L_v\end{bmatrix}\\
\notag=&0.5\rho_{\text{air}}V^2_{\text{w}}\text{diag}\left[\cos(\eta_{\psi}-\beta_w)A_{T},\sin(\eta_{\psi}-\beta_w)A_{L},\right.\\
&\left. \sin[2(\eta_{\psi}-\beta_w)]A_{L}L_v\right]\begin{bmatrix}C_x\\C_y\\C_N\end{bmatrix}=\Pi\Phi
\end{align}
where,
\begin{align}
\hspace{-5mm}
\label{windmod}
\notag\Pi=&0.5\rho_{\text{air}}V^2_{\text{w}}\text{diag}\left[\cos(\eta_{\psi}-\beta_w)A_{T},\sin(\eta_{\psi}-\beta_w)A_{L},\right.\\
&\left. \sin[2(\eta_{\psi}-\beta_w)]A_{L}L_v\right], \Phi=\begin{bmatrix}C_x\\C_y\\C_N\end{bmatrix}
\end{align}
$\rho_{\text{air}}$ is the density of air. $\Phi$ is the peak value of wind drag coefficient. $A_T$, $A_L$ and $L_v$ denote the transverse projected area, lateral projected area and the length of the vessel. $\beta_w$ represents the attack direction of the wind. $V_w$ is the relative velocity between the wind and the vessel. Next, we present some assumptions and remarks to facilitate the sequent development.
\begin{assumption}
\label{ass1}
The inertia matrix $M$ is invertible and $M^{-1}$ is bounded. The upper bound can be expressed as $||M^{-1}||_{\infty}\leq \overline{M^{-1}}$, where $\overline{M^{-1}} \in \mathbb{R}$ is a positive constant bound.
\end{assumption}
\begin{assumption}
\label{ass2}
The disturbance term $d$ is bounded with $d^Td\leq\overline{d}$. $\overline{d}\in \mathbb{R}$ is a positive constant.
\end{assumption}
\begin{assumption}
\label{ass3}
In this paper, we only consider the drift wave-induced force and moment which is a low-frequency part of the wave effects. The high-frequency part is ignored.
\end{assumption}
\begin{remark}
Assumption \ref{ass3} implies that there is no need to enclose a filter on the position and velocity signal, $\eta$ and $\nu$, during the control design.
\end{remark}
\section{Adaptive Sea State Observer}
In this brief, a novel sea state observer is built to alarm the shielding effect as well as approximate the wind force and moment. To achieve this, the idea of fault detection and diagnosis is incorporated by building a model-based nonlinear observer with full state feedback. The wave-induced drift force under the shielding effect can be regarded as an evolutive fault. Large wave-induced force can be alarmed by investigating the output of the wind estimator and the residual error of the observer. In this paper, the wind and wind-generated wave force are both assumed to propagate along the $X_0$ direction. Initially, due to the shielding effect, the vessel is subject to the weak wind force solely. A wind drag coefficient estimator is developed to adaptively estimate the unknown peak value of wind drag coefficient $\Phi$. When the shadow influence vanishes, the estimator would fall into overcompensation and the observation error increase. These phenomenon help us to judge the occurrence of large wave-induced force. Then, NNs which have inherent approximation capabilities \cite{ge2003neural} \cite{ge2013stable} are applied in sea observer and the controller to encounter the uncertain wave force. The design of sea state observer is introduced in this section.\\
\indent The more complicated observer after alarm with NN compensation is presented first. The formulation in (\ref{overallmodel1}) (\ref{overallmodel2}) can be rewritten into a more compact form as
\be
\dot{X}=f(X)X+\phi(X)+R[\tau(t-t_d)+\gamma(t-T)\tau_{\text{wave}}+\tau_{\text{wind}}+d]
\ee
where $X=[\eta^T,\nu^T]^T$, $f(X)=\begin{bmatrix}O &J(\eta_{\psi})\\O&-M^{-1}[C(\nu)+D(\nu)]\end{bmatrix}$, $\phi(X)=\begin{bmatrix}O\\-M^{-1}g(\eta)\end{bmatrix}$, $R=\begin{bmatrix}O\\M^{-1}\end{bmatrix}$.
Add and minus $AX$ at the right hand side of the above expression, we obtain.
\begin{align}
\label{demod}
\notag\dot{X}&=AX+[f(X)-A]X+\phi(X)\\
&+R\left[\tau(t-t_d)+\gamma(t-T)\tau_{\text{wave}}+\tau_{\text{wind}}+d\right]
\end{align}
where $A^T=A \in \mathbb{R}^{6\times6}$ matrix is chosen to be Hurwitz and the pair $(A,R)$ is completely controllable. According to Kalman-Yakubovich-Popov (KYP) lemma \cite{kalman1963lyapunov}, there exists a symmetric matrix $P$ and a vector $Q$ satisfying
\bea
\label{8}
A^TP+PA=-QQ^T
\eea
\begin{assumption}
\label{lips}
$[f(X)-A]X+\phi(X)$ is Lipschitz and satisfies $\big\|[f(X_1)-A]X_1+\phi(X_1)-[f(X_2)-A]X_2-\phi(X_2)\big\|\leq \sigma_{d}||X_1-X_2||$ where $\sigma_{d}$ is Lipschitz constant.
\end{assumption}
\indent A set of linearly parameterized NNs with Radial Basis Function (RBF) \cite{ge1998adaptive} is employed to handle the unknown wave force.\\
Consider
\bea
\gamma(t-T)\tau_{\text{wave}}(Z_{ow})= W_d^{*T}S(Z_{ow})+\epsilon
\eea
with
\bea
\gamma(t-T)\hat{\tau}_{\text{wave}}(\hat{Z}_{ow})= \hat{W}_d^{T}S(\hat{Z}_{ow})
\eea
we can further obtain
\begin{align}
\notag&\gamma(t-T)\tau_{\text{wave}}(Z_{ow})-\gamma(t-T)\hat{\tau}_{\text{wave}}(\hat{Z}_{ow})\\
\notag=&W_d^{*T}S(Z_{ow})-W_d^{*T}S(\hat{Z}_{ow})+W_d^{*T}S(\hat{Z}_{ow})\\
\notag&-\hat{W}_d^{T}S(\hat{Z}_{ow})+\epsilon\\
\notag=&\tilde{W}_d^TS(\hat{Z}_{ow})+W^{*T}_d\big[S(Z_{ow})-S(\hat{Z}_{ow})\big]+\epsilon\\
=&\tilde{W}_d^TS(\hat{Z}_{ow})+\Lambda
\end{align}
where $\hat{W}_d=\text{blockdiag}\left[\hat{W}_{\text{d1}},\hat{W}_{\text{d2}},...,\hat{W}_{\text{d6}} \right]$ is the weight matrix. $W^*_d$ is the corresponding optimal weights and define $\tilde{W}_d=\hat{W}_d-W_d^{*}$. The input of the network is $Z_{ow}=[P^T_{\text{wave}},X^T]^T$. $P^T_{\text{wave}}$ is the wave-related measured parameters. Since the activation function is bounded, $S(Z_{ow})-S(\hat{Z}_{ow})$ is bounded. Moreover, $W^*_d$ and the approximation error $\epsilon$ are bounded, hence, the newly defined disturbance term $\Lambda=W^{*T}_d\big[S(Z_{ow})-S(\hat{Z}_{ow})\big]+\epsilon$ is bounded, and it satisfies
\bea
\label{14}
||\Lambda||^2\leq\overline{\Lambda}
\eea
where $\overline{\Lambda} \in \mathbb{R}$ is the constant upper bound. The observer after alarm is designed to be
\begin{align}
\label{sevdet}
\notag\dot{\hat{X}}=&A\hat{X}+\big[f(\hat{X})-A\big]\hat{X}+\phi(\hat{X})+R\big[\tau(t-t_d)\\
\notag&+\gamma(t-T)\tau_{\text{wave}}+\hat{\tau}_{\text{wind}}\big]+L\big[CX-C\hat{X}\big]\\
\notag=&f(\hat{X})\hat{X}+\phi(\hat{X})+R\big[\tau(t-t_d)+\hat{W}_d^TS(\hat{Z}_{ow})\\
&+\hat{\tau}_{\text{wind}}\big]+L\big[CX-C\hat{X}\big]
\end{align}
where $\hat{X}$ is the estimation of $X$. $L=P^{-1}C^T \in \mathbb{R}^{6\times6}$ is a observer gain matrix. $C\in\mathbb{R}^{6\times6}$ is the measurement matrix. $\hat{\tau}_{\text{wind}}$ denotes the wind force estimator to be developed later. Define the observer error as $\tilde{X}=X-\hat{X}$. The derivative of $\tilde{X}$ is
\begin{align}
\hspace{-5mm}
\notag\dot{\tilde{X}}=&\dot{X}-\dot{\hat{X}}\\
\notag=&(A-LC)\tilde{X}+\big[(f(X)-A)X+\phi(X)-(f(\hat{X})\\
\notag&-A)\hat{X}-\phi(\hat{X})\big]+R\big[\tau_{\text{wind}}-\hat{\tau}_{\text{wind}}+\tilde{W}_d^TS(\hat{Z}_{ow})\\
&+\Lambda+d\big]
\label{asdf}
\end{align}
For stability analysis of error signals, the following Lyapunov candidate is considered
\be
V=\tilde{X}^TP\tilde{X}+\displaystyle\frac{1}{2}\tilde{\Phi}^T\Gamma^{-1}\tilde{\Phi}+\sum^6_{i=1}\displaystyle\frac{1}{\omega_i}\tilde{W}_{di}^T\tilde{W}_{di}
\ee
where $\omega_i, (i=1,2,...,6)$ is a constant value. The error of wind coefficient estimation $\tilde{\Phi}$ is
\bea
\tilde{\Phi}=\Phi-\hat{\Phi}
\eea
Incorporating (\ref{asdf}), the time derivative of $V$ gives
\begin{align}
\hspace{-5mm}
\notag\dot{V}=&2\tilde{X}^TP\dot{\tilde{X}}+\dot{\tilde{\Phi}}^T\Gamma^{-1}\tilde{\Phi}+\sum^n_{i=1}\displaystyle\frac{2}{\omega_i}\tilde{W}_{di}^T\dot{\tilde{W}}_{di}\\
\notag=&2\tilde{X}^TP\big[(A-LC)\tilde{X}+[(f(X)-A)X+\phi(X)\\
\notag&-(f(\hat{X})-A)\hat{X}-\phi(\hat{X})]+R(\tau_{\text{wind}}-\hat{\tau}_{\text{wind}}+\tilde{W}_d^T\\
&S(\hat{Z}_{ow})+\Lambda+d)\big]+\dot{\tilde{\Phi}}^T\Gamma^{-1}\tilde{\Phi}+\sum^n_{i=1}\displaystyle\frac{2}{\omega_i}\tilde{W}_{di}^T\dot{\tilde{W}}_{di}
\end{align}
Consider Assumption \ref{lips}, $\dot{V}$ becomes
\begin{align}
\hspace{-5mm}
\label{V1}
\notag\dot{V}\leq &2\tilde{X}^TP\big[(A-LC)\tilde{X}+R(\tau_{\text{wind}}-\hat{\tau}_{\text{wind}}+\tilde{W}_d^T\\
\notag&S(\hat{Z}_{ow})+\Lambda+d)\big]+2\sigma\|P\tilde{X}\| \|\tilde{X}\|+\dot{\tilde{\Phi}}^T\Gamma^{-1}\tilde{\Phi}\\
\notag&+\sum^n_{i=1}\displaystyle\frac{2}{\omega_i}\tilde{W}_{di}^T\dot{\tilde{W}}_{di}\\
\notag=&2\tilde{X}^TP(A-LC)\tilde{X}+2\tilde{X}^TPR(\Pi\Phi-\hat{\tau}_{\text{wind}}\\
\notag&+\tilde{W}_d^TS(\hat{Z}_{ow})+\Lambda+d)+2\sigma\|P\tilde{X}\| \|\tilde{X}\|\\
&+\dot{\tilde{\Phi}}^T\Gamma^{-1}\tilde{\Phi}+\sum^n_{i=1}\displaystyle\frac{2}{\omega_i}\tilde{W}_{di}^T\dot{\tilde{W}}_{di}
\end{align}
The adaptive law of $\hat{\Phi}$ is designed as
\bea
\label{adawind}
\dot{\hat{\Phi}}=2\Gamma^T\Pi^TR^TP^T\tilde{X}
\eea
With the adaptive law above, we have
\bea
\label{dphi}
\dot{\tilde{\Phi}}^T\Gamma^{-1}\tilde{\Phi}=-2\tilde{X}^TPR\Pi\tilde{\Phi}
\eea
Consequently, the wind force estimation term $\hat{\tau}_{\text{wind}}$ can be calculated as
\bea
\label{mu}
\hat{\tau}_{\text{wind}}=\Pi\hat{\Phi}
\eea
Substituting (\ref{dphi}) and (\ref{mu}) into (\ref{V1}), we obtain
\begin{align}
\hspace{-15mm}
\label{21}
\notag\dot{V} \leq &2\tilde{X}^TP(A-LC)\tilde{X}+2\tilde{X}^TPR\big(\Pi\Phi-\Pi\hat{\Phi}+\tilde{W}_d^T\\
\notag&S(\hat{Z}_{ow})+\Lambda+d\big)+2\sigma_d\|P\tilde{X}\| \|\tilde{X}\|-2\tilde{X}^TPR\Pi\tilde{\Phi}\\
\notag&+\sum^n_{i=1}\displaystyle\frac{2}{\omega_i}\tilde{W}_{di}^T\dot{\tilde{W}}_{di}\\
\notag=&2\tilde{X}^TP(A-LC)\tilde{X}+2\tilde{X}^TPR\big(\tilde{W}_d^TS(\hat{Z}_{ow})+\Lambda\\
&+d\big)+2\sigma_d\|P\tilde{X}\| \|\tilde{X}\|+\sum^n_{i=1}\displaystyle\frac{2}{\omega_i}\tilde{W}_{di}^T\dot{\tilde{W}}_{di}
\end{align}
Designing the adaptation for the weights in NN as
\bea
\label{nndet}
\dot{\hat{W}}_{di}=-\omega_i(\tilde{X}^TPR)_i S(\hat{Z}_{ow})
\eea
where $(\bullet)_i, (i=1,2...,6)$ is the $i$th column of $\bullet$. Invoking the update law into (\ref{21}), we further have
\be
\label{23}
\dot{V}\leq 2\tilde{X}^TP(A-LC)\tilde{X}+2\tilde{X}^TPR(\Lambda+d)+2\sigma_d\|P\tilde{X}\| \|\tilde{X}\|
\ee
\begin{lemma}
\label{lemma1}
\cite{chen2010robust} For any two matrices $X_{l1}$ and $Y_{l1}$ of the same dimension, there exists a positive constant $c_{l1}$ such that the following inequality holds.
\bea
X_{l1}^TY_{l1}+Y_{l1}^TX_{l1}\leq c_{l1}X_{l1}^TX_{l1}+c_{l1}^{-1} Y_{l1}^TY_{l1}
\eea
\end{lemma}
Since $2\tilde{X}^TPR(\Lambda+d)$ is a scalar and considering Lemma \ref{lemma1}, Assumption \ref{ass2} and (\ref{14}), we have the following inequalities.
\begin{align}
\hspace{-5mm}
\label{24}
\notag&2\tilde{X}^TPR\Lambda=\tilde{X}^TPR\Lambda+\Lambda^TR^TP^T\tilde{X}\leq \kappa_1\tilde{X}^TPRR^T\\
&P^T\tilde{X}+\kappa_1^{-1}\Lambda^T\Lambda \leq \kappa_1\tilde{X}^TPRR^TP^T\tilde{X}+\kappa_1^{-1}\overline{\Lambda}
\end{align}
\begin{align}
\hspace{-5mm}
\label{25}
\notag&2\tilde{X}^TPRd=\tilde{X}^TPRd+d^TR^TP^T\tilde{X}\leq \kappa_2\tilde{X}^TPRR^T\\
&P^T\tilde{X}+\kappa_2^{-1}d^Td \leq \kappa_2\tilde{X}^TPRR^TP^T\tilde{X}+\kappa_2^{-1}\overline{d}
\end{align}
Moreover, it is clear that the following fact is held:
\begin{align}
\hspace{-5mm}
\label{26}
\notag2\sigma_d\|P\tilde{X}\| \|\tilde{X}\|&\leq 2\sigma_d\|\lambda_{\text{max}}(P)\tilde{X}\| \|\tilde{X}\|\\
&=2\sigma_d\lambda_{\text{max}}(P)\|\tilde{X}\|^2=\tilde{X}^T2\sigma_d\lambda_{\text{max}}(P)I\tilde{X}
\end{align}
where $\lambda_{\text{max}}(\bullet)$ is the maximum eigenvalue of $\bullet$.
Substituting (\ref{24}) (\ref{25}) and (\ref{26})into (\ref{23}) yields
\begin{align}
\label{31}
\hspace{-5mm}
\notag\dot{V}\leq& \tilde{X}^T(2PA-2PLC+2\sigma_d\lambda_{\text{max}}(P)I+\kappa_1PRR^TP^T\\
&+\kappa_2PRR^TP^T)\tilde{X}+\kappa_1^{-1}\overline{\Lambda}+\kappa_2^{-1}\overline{d}
\end{align}
In accordance with $L=P^{-1}C^T$ and KYP lemma, (\ref{31}) gives
\begin{align}
\hspace{-5mm}
\notag\dot{V} \leq& \tilde{X}(A^TP+PA-2C^TC+2\sigma_d\lambda_{\text{max}}(P)I+\kappa_1PR\\
\notag&R^TP^T+\kappa_2PRR^TP^T)\tilde{X}+\kappa_1^{-1}\overline{\Lambda}+\kappa_2^{-1}\overline{d}\\
\notag=&\tilde{X}^T(-QQ^T-2C^TC+2\sigma_d\lambda_{\text{max}}(P)I+\kappa_1PRR^TP^T\\
\notag&+\kappa_2PRR^TP^T)\tilde{X}+\kappa_1^{-1}\overline{\Lambda}+\kappa_2^{-1}\overline{d}\\
=&\tilde{X}^TE\tilde{X}+\kappa_1^{-1}\overline{\Lambda}+\kappa_2^{-1}\overline{d}
\end{align}
where $E=-QQ^T-2C^TC+2\sigma_d\lambda_{\text{max}}(P)I+(\kappa_1+\kappa_2)PRR^TP^T$. By properly choosing $A$, $P$, $Q$, $\sigma_d$, $\kappa_1$ and $\kappa_2$, $E$ can be guaranteed to be negative definite and $\tilde{X}^TE\tilde{X}<0$.\\
If
\be
-\tilde{X}^TE\tilde{X}=\tilde{X}^T(-E)\tilde{X} \geq \lambda_{\text{min}}(-E)\|\tilde{X}\|^2>\kappa_1^{-1}\overline{\Lambda}+\kappa_2^{-1}\overline{d}
\ee
we can ensure $\dot{V}<0$. The stability condition above can be further expressed as
\bea
\|\tilde{X}\|>\sqrt{\displaystyle\frac{\kappa_1^{-1}\overline{\Lambda}+\kappa_2^{-1}\overline{d}}{\lambda_{\text{min}}(-E)}}
\eea
\begin{remark}
By proper selection of the observer coefficients, the estimation error, i.e. $\tilde{X}$ can be arbitrarily small.
\end{remark}
\indent Since only wind-induced forces and moment affecting the motion of the vessel before the vessel is subject to large wave-induced force, the wave force term $\gamma(t-T)\tau_{\text{wave}}$ in (\ref{demod}) can be ignored. The sea observer under this stage is proposed in the following pattern.
\begin{align}
\hspace{-5mm}
\label{besever}
\dot{\hat{X}}=f(\hat{X})\hat{X}+\phi(\hat{X})+R\big[\tau(t-t_d)+\Pi\hat{\Phi}\big]+L[CX-C\hat{X}]
\end{align}
\begin{remark}
The stability verification is very similar to the observer with NN estimator above thus is neglected. In practical use, when the sea state changes, the wind force estimator will overly compensate due to the involvement of the wave force. Therefore, we can judge the moment of alarm by monitoring the estimated wind drag coefficients. The NN compensator in both observer and controller are to be activated when a designed threshold for estimated wind drag coefficients are exceeded. The observer error $\tilde{X}$ can also be applied as an axillary indicator for the alarm.
\end{remark}
\begin{remark}
Based on the \emph{Helmholtz-Kirchhoff} plate theory \cite{fossen2011handbook}, the peak of wind drag coefficient is parameterized in terms of four shape-related parameters. Hence, for fixed vessel, the alarm threshold is unique and can be calculated approximately or through field calibration.
\end{remark}
\section{Robust Control Design}
In this section, we focus on an input time delay control with constrained tracking error. One approach to cope with the input time delay is to convert the original system into a delay-free system known as the Artstein model \cite{1103023}. Essentially, Artstein model is a predictor-like controller for linear system. However, the dynamics of the vessel is of great nonlinearity and this model does not consider the limitation of tracking error. Therefore, inspired by \cite{1103023} and combining BLF method \cite{tee2009barrier}, a model-based robust controller with input time delay and tracking error constraint is developed in this paper.
\subsection{Design of control before alarm}
The wind force is estimated using $\hat{\tau}_{\text{wind}}$ in the last section. Define the estimation error as $\tilde{\tau}_{\text{wind}}=\tau_{\text{wind}}-\hat{\tau}_{\text{wind}}$.
When no large wave-induced drift force is detected, we consider the following dynamic system.
\be
\label{modelconbe}
M\dot{\nu}+C(\nu)\nu+D(\nu)\nu+g(\eta)=\tau^{'}(t-t_d)+d_1
\ee
where $\tau^{'}(t-t_d)=\tau(t-t_d)+\hat{\tau}_{\text{wind}}$, $d_1=\tilde{\tau}_{\text{wind}}+d$ which performs as a feedforward control to cope with the wind force. While, the actuator delay of the feedforward control component $\hat{\tau}_{\text{wind}}$ is neglected in this work. The input delay $t_d$ is assumed as a known constant value.
\begin{remark}
The estimation error of the peak of wind drag coefficient $\tilde{\Phi}$ has been proven to be bounded in the last section. Hence, the wind force estimation error $\tilde{\tau}_{\text{wind}}$ is bounded. Combining Assumption \ref{ass2}, the newly defined term $d_1$ is bounded and can be rationally limited as $\overline{d_1}$ with $||d_1||\leq\overline{d_1}$. Where $\overline{d_1}$ is a positive constant.
\end{remark}
\indent Incorporating Symmetry Barrier Lyapunov Function (SBLF) \cite{tee2009barrier}, a backstepping approach is employed to design the control.\\ \emph{Step 1}: Denote
\bea
\label{37}
z_1=\eta_d-\eta, \quad z_2=\alpha_c-\nu
\eea
where the desired trajectory satisfies $\eta_d, \dot{\eta}_d \in \mathscr{L}_{\infty}$. $\alpha_c$ is the stabilizing function. Choose a positive definite and $C^{1}$ continuous SBLF candidate as
\begin{align}
\hspace{-5mm}
\notag V_1=&\displaystyle\frac{1}{2}\text{log}\displaystyle\frac{N^T_{b}I_xN_{b}}{N^T_{b}I_xN_{b}-z^T_1I_xz_1}+\displaystyle\frac{1}{2}\text{log}\displaystyle\frac{N^T_{b}I_yN_{b}}{N^T_{b}I_yN_{b}-z^T_1I_yz_1}\\
&+\displaystyle\frac{1}{2}\text{log}\displaystyle\frac{N^T_{b}I_{\psi}N_{b}}{N^T_{b}I_{\psi}N_{b}-z^T_1I_{\psi}z_1}
\end{align}
where
\bea
I_x=\begin{bmatrix} 1&0&0\\0&0&0\\0&0&0\end{bmatrix} \quad I_y=\begin{bmatrix} 0&0&0\\0&1&0\\0&0&0\end{bmatrix} \quad I_{\psi}=\begin{bmatrix} 0&0&0\\0&1&0\\0&0&0\end{bmatrix}
\eea
$N_b \in \mathbb{R}^{3\times1}$ is the tracking error constraint such that $|z_1|\leq N_b$ should be satisfied.
\begin{remark}
\label{rem5}
In practical use, the initial condition of position and velocity of the vessel are consistent with the desired trajectory. Hence, $|z_1(0)|<N_b$ can be guaranteed.
\end{remark}
Time derivative of $V_1$ yields
\begin{align}
\hspace{-5mm}
\label{dV}
\notag \dot{V}_1=&\displaystyle\frac{z_1^TI_x\dot{z}_1}{N^T_{b}I_xN_{b}-z^T_1I_xz_1}+\displaystyle\frac{z_1^TI_y\dot{z}_1}{N^T_{b}I_yN_{b}-z^T_1I_yz_1}\\
&+\displaystyle\frac{z_1^TI_{\psi}\dot{z}_1}{N^T_{b}I_{\psi}N_{b}-z^T_1I_{\psi}z_1}
\end{align}
Differentiating $z_1$ with respect to time gives
\bea
\label{dz}
\dot{z}_1=\dot{\eta}_d-J(\eta_{\psi})(\alpha_c-z_2)
\eea
Substituting (\ref{dz}) into (\ref{dV}), we have
\begin{align}
\hspace{-5mm}
\label{dV2}
\notag &\dot{V}_1=\displaystyle\frac{z_1^TI_x\left[\dot{\eta}_d-J(\eta_{\psi})(\alpha_c-z_2)\right]}{N^T_{b}I_xN_{b}-z^T_1I_xz_1}+\displaystyle\frac{z_1^TI_y\left[\dot{\eta}_d-J(\eta_{\psi})\right.}{N^T_{b}I_yN_{b}}\\
&\displaystyle\frac{\left.(\alpha_c-z_2)\right]}{-z^T_1I_yz_1}+\displaystyle\frac{z_1^TI_{\psi}\left[\dot{\eta}_d-J(\eta_{\psi})(\alpha_c-z_2)\right]}{N^T_{b}I_{\psi}N_{b}-z^T_1I_{\psi}z_1}
\end{align}
Design the stabling function $\alpha_c$ to be
\bea
\label{al1}
\alpha_c=J^T(\eta_{\psi})[\dot{\eta}_d+(N_b^TN_b-z_1^Tz_1)K_1z_1]
\eea
Substituting (\ref{al1}) into (\ref{dV2}) and considering the property of rotation matrix $J(\eta_{\psi})J^T(\eta_{\psi})=I$, following equation is achieved.
\begin{align}
\label{455}
\hspace{-5mm}
\notag \dot{V}_1=&-3z_1^TK_1z_1+\displaystyle\frac{z_1^TI_xJ(\eta_{\psi})z_2}{N^T_{b}I_xN_{b}-z^T_1I_xz_1}\\
&+\displaystyle\frac{z_1^TI_yJ(\eta_{\psi})z_2}{N^T_{b}I_yN_{b}-z^T_1I_yz_1}+\displaystyle\frac{z_1^TI_{\psi}J(\eta_{\psi})z_2}{N^T_{b}I_{\psi}N_{b}-z^T_1I_{\psi}z_1}
\end{align}
\emph{Step 2}:
Define an auxiliary state $S \in \mathbb{R}^{3 \times 1}$ to compensate for the input delay with the following expression.
\bea
\label{S1}
S=z_2-M^{-1}\int^{t}_{t-t_d}\tau^{'}(\theta)d\theta-z_f
\eea
where $z_f \in \mathbb{R}^{3 \times 1}$ satisfies the following adaptive law.
\bea
\label{42}
\dot{z}_f=K_2S-\Gamma_1z_2-\Theta z_f
\eea
In (\ref{42}), $K_2, \Gamma_1, \Theta \in \mathbb{R}^{3 \times 3}$ are positive tuning parameters. Multiply both sides of (\ref{S1}) by $M$ and denote $M_s=C(\nu)+D(\nu)+g(\eta)$, the derivative of $MS$ yields
\begin{align}
\hspace{-5mm}
\label{44}
\notag M\dot{S}=&M\dot{z}_2-\tau^{'}(t)+\tau^{'}(t-t_d)-\dot{z}_f\\
\notag=&M\dot{\alpha}_c+C(\nu)\nu+D(\nu)\nu+g(\eta)-d_1-\tau^{'}(t)\\
\notag&-K_2S+\Theta z_f+\Gamma_1z_2\\
\notag=&M\dot{\alpha}_c+M_s-d_1+N_c-\tau^{'}(t)-K_2S-K_2z_2\\
&-(S^{T})^{+}\dot{S}^Tz_2
\end{align}
where $N_c$ is defined as follows and consider the Mean Value Theorem \cite{de1997adaptive}.
\begin{align}
\hspace{-5mm}
\label{45}
\notag &N_c=\Theta z_f+\Gamma_1 z_2+K_2z_2+(S^{T})^{+}\dot{S}^Tz_2\\
&||N_c||\leq \overline{N}_c(||z_{s}||)||z_{s}||
\end{align}
where the bounding function $\overline{N}_c(||z_{s}||)$ is a globally positive function. $z_{s}$ has the definition of $z_{s}=[z^T_1, z^T_2, S^T, z^T_{\tau}, z^T_f]^T$, where $z_{\tau} \in \mathbb{R}^{3\times1}$ denotes
\bea
z_{\tau}=\tau^{'}(t)-\tau^{'}(t-\beta)=\int^t_{t-t_d}\dot{\tau}^{'}(\theta)d\theta
\eea
With the involvement of auxiliary state $S$, the delayed system is converted into a delay-free one as shown in (\ref{44}). For the velocity of the vessel, no limitation is needed. Thus, a quadratic form Lyapunov-Krasovskii candidate function is defined as \cite{mazenc2006backstepping}
\begin{align}
\hspace{-5mm}
\notag V_2=&V_1+\displaystyle\frac{1}{2}z_2^Tz_2+\displaystyle\frac{1}{2}S^TMS+\displaystyle\frac{1}{2}z^T_f z_f\\
&+\upsilon\int^t_{t-t_d}(\int^t_{w}||\dot{\tau}^{'}(\theta)||^2d\theta) dw
\end{align}

Differentiating $V_2$ and invoking (\ref{455}), (\ref{S1}), (\ref{42}) and (\ref{44}), we obtain
\begin{align}
\hspace{-5mm}
\label{48}
\notag\dot{V}_2=&\dot{V}_1+z_2^T\dot{z}_2+S^TM\dot{S}+z_f\dot{z}_f+\upsilon t_d||\dot{\tau}^{'}(\theta)||^2\\
\notag &-\upsilon\int^t_{t-t_d}||\dot{\tau}^{'}(\theta)||^2d\theta\\
\notag=&-3z_1^TK_1z_1+\displaystyle\frac{z_1^TI_xJ(\eta_{\psi})z_2}{N^T_{b}I_xN_{b}-z^T_1I_xz_1}\\
\notag&+\displaystyle\frac{z_1^TI_yJ(\eta_{\psi})z_2}{N^T_{b}I_yN_{b}-z^T_1I_yz_1}+\displaystyle\frac{z_1^TI_{\psi}J(\eta_{\psi})z_2}{N^T_{b}I_{\psi}N_{b}-z^T_1I_{\psi}z_1}\\
\notag&+z_2^T(\dot{S}-M^{-1}(\tau(t-t_d)-\tau(t))+K_2 S-\Theta z_f\\
\notag&-\Gamma_1z_2)+S^T(M\dot{\alpha}_c+M_s-d_1+N_c-\tau(t)^{'}\\
\notag&-K_2S-K_2z_2-(S^{T})^{+}\dot{S}^Tz_2)+z^T_fK_2S-z^T_f \Theta z_f\\
\notag&-z_f^T\Gamma_1z_2+\upsilon t_d||\dot{\tau}^{'}(\theta)||^2-\upsilon\int^t_{t-t_d}||\dot{\tau}^{'}(\theta)||^2d\theta\\
\notag=&-3z_1^TK_1z_1+\displaystyle\frac{z_1^TI_xJ(\eta_{\psi})z_2}{N^T_{b}I_xN_{b}-z^T_1I_xz_1}\\
\notag&+\displaystyle\frac{z_1^TI_yJ(\eta_{\psi})z_2}{N^T_{b}I_yN_{b}-z^T_1I_yz_1}+\displaystyle\frac{z_1^TI_{\psi}J(\eta_{\psi})z_2}{N^T_{b}I_{\psi}N_{b}-z^T_1I_{\psi}z_1}\\
\notag&-z_2^T\Gamma_1z_2+z_2^TM^{-1}z_{\tau}-z_2^T(\Gamma_1+I)z_f-S^T K_2 S\\
\notag&+S^T\big[M\dot{\alpha}_c+M_s-d_1+N_c-\tau^{'}(t)\big]+z_f^TK_2S\\
&-z_f^T \Theta z_f+\upsilon t_d||\dot{\tau}^{'}(\theta)||^2-\upsilon\int^t_{t-t_d}||\dot{\tau}^{'}(\theta)||^2d\theta
\end{align}
Design the following control law
\begin{align}
\hspace{-5mm}
\label{tau11}
\notag\tau^{'}(t)=&M\dot{\alpha}_c+M_s+K_2z_f+(S^T)^{+}\bigg[\displaystyle\frac{z_1^TI_xJ(\eta_{\psi})z_2}{N^T_{b}I_xN_{b}-z^T_1I_xz_1}\\
\notag&+\displaystyle\frac{z_1^TI_yJ(\eta_{\psi})z_2}{N^T_{b}I_yN_{b}-z^T_1I_yz_1}+\displaystyle\frac{z_1^TI_{\psi}J(\eta_{\psi})z_2}{N^T_{b}I_{\psi}N_{b}-z^T_1I_{\psi}z_1}\\
\notag&+\displaystyle\frac{N^T_b I_x N_b z^T_1 K_1 z_1}{N^T_{b}I_{x}N_{b}-z^T_1I_{x}z_1}+\displaystyle\frac{N^T_b I_y N_b z^T_1 K_1 z_1}{N^T_{b}I_{y}N_{b}-z^T_1I_{y}z_1}\\
&+\displaystyle\frac{N^T_b I_{\psi} N_b z^T_1 K_1 z_1}{N^T_{b}I_{\psi}N_{b}-z^T_1I_{\psi}z_1}\bigg]
\end{align}
Substitute (\ref{tau11}) into (\ref{48}) and considering (\ref{45}) and Assumption \ref{ass1}, we have
\begin{align}
\hspace{-5mm}
\label{51}
\notag\dot{V}_2=&-3z_1^TK_1z_1-z_2^T\Gamma_1z_2-S^T K_2S-z_f^T \Theta z_f+z_2^TM^{-1}\\
\notag&z_{\tau}-z_2^T(\Gamma_1+I)z_f+S^TN_c-S^Td_1+\upsilon t_d||\dot{\tau}^{'}(\theta)||^2\\
\notag&-\upsilon\int^t_{t-t_d}||\dot{\tau}^{'}(\theta)||^2d\theta-\displaystyle\frac{N^T_b I_x N_b z^T_1 K_1 z_1}{N^T_{b}I_{x}N_{b}-z^T_1I_{x}z_1}\\
\notag&-\displaystyle\frac{N^T_b I_y N_b z^T_1 K_1 z_1}{N^T_{b}I_{y}N_{b}-z^T_1I_{y}z_1}-\displaystyle\frac{N^T_b I_{\psi} N_b z^T_1 K_1 z_1}{N^T_{b}I_{\psi}N_{b}-z^T_1I_{\psi}z_1}\\
\notag\leq & -3z^T_1K_1z_1-\lambda_{\text{min}}(\Gamma_1)z^T_2z_2-S^TK_2S\\
\notag &-\lambda_{\text{min}}(\Theta_1)z_f^Tz_f+\overline{M^{-1}}||z_2||||z_{\tau}||+\overline{(-\Gamma_1-I)}\\
\notag &||z_2||||z_f||+\overline{N}_c(||z_s||)||z_s||||S||+\overline{d_1}||S||+\upsilon t_d\\
\notag&||\dot{\tau}^{'}(\theta)||^2-\upsilon\int^t_{t-t_d}||\dot{\tau}^{'}(\theta)||^2d\theta-\displaystyle\frac{N^T_b I_x N_b z^T_1 K_1 z_1}{N^T_{b}I_{x}N_{b}-z^T_1I_{x}z_1}\\
&-\displaystyle\frac{N^T_b I_y N_b z^T_1 K_1 z_1}{N^T_{b}I_{y}N_{b}-z^T_1I_{y}z_1}-\displaystyle\frac{N^T_b I_{\psi} N_b z^T_1 K_1 z_1}{N^T_{b}I_{\psi}N_{b}-z^T_1I_{\psi}z_1}
\end{align}
To facilitate the subsequent analysis, the Young's inequality is introduced.
\bea
\label{55}
||a||||b||\leq \displaystyle\frac{\iota}{4}||a||^2+\displaystyle\frac{1}{\iota}||b||^2
\eea
where $a$ and $b$ are vectors, $\iota$ is a positive constant. Therefore, the $\overline{N}_c(||z_s||)||z_s||||S||$ term in (\ref{51}) yields
\begin{align}
\label{56}
\hspace{-5mm}
\notag&\overline{N}_c(||z_s||)||z_s||||S||\leq \displaystyle\frac{\sigma_3}{4}\overline{N}^2_c(||z_s||)||z_s||^2+\displaystyle\frac{1}{\sigma_3}||S||^2\\
\notag&\leq\displaystyle\frac{\sigma_3}{4}\overline{N}^2_c\big(||z_s||)(||z_1||^2+||z_2||^2+||S||^2+||z_{\tau}||^2\\
&+||z_f||^2\big)+\displaystyle\frac{1}{\sigma_3}||S||^2
\end{align}
Similar situation holds for other terms in (\ref{51}). Moreover, under the condition of $||z_1||<||N_b||$, the following inequalities holds.
\begin{align}
\label{57}
\hspace{-5mm}
\notag&\displaystyle\frac{\sigma_3}{12}\overline{N_c}^2(||z_s||)z_1^Tz_1-z_1^TK_1z_1-\displaystyle\frac{N^T_b I_x N_b z^T_1 K_1 z_1}{N^T_{b}I_{x}N_{b}-z^T_1I_{x}z_1}\\
\notag&\leq -\displaystyle\frac{(\lambda_{\text{min}}(K_1)-\displaystyle\frac{\sigma_3}{12}\overline{N_c}^2(||z_s||))N_b^TI_xN_bz_1^TI_xz_1}{N^T_{b}I_{x}N_{b}-z^T_1I_{x}z_1}\\
\notag&\leq -\left(\lambda_{\text{min}}(K_1)-\displaystyle\frac{\sigma_3}{12}\overline{N_c}^2(||z_s||)\right)N_b^TI_xN_b\\
&\text{log}\displaystyle\frac{N^T_{b}I_{x}N_{b}}{N^T_{b}I_{x}N_{b}-z^T_1I_{x}z_1}
\end{align}
For $y$ and $\psi$, we have identical transformation. Herein, define
\begin{align}
\label{58}
\hspace{-5mm}
\notag \Xi_x=&-\left(\lambda_{\text{min}}(K_1)-\displaystyle\frac{\sigma_3}{12}\overline{N_c}^2\left(||z_s||\right)\right)N_b^TI_xN_b\\
\notag&\text{log}\displaystyle\frac{N^T_{b}I_{x}N_{b}}{N^T_{b}I_{x}N_{b}-z^T_1I_{x}z_1}\\
\notag \Xi_y=&-\left(\lambda_{\text{min}}(K_1)-\displaystyle\frac{\sigma_3}{12}\overline{N_c}^2\left(||z_s||\right)\right)N_b^TI_yN_b\\
\notag&\text{log}\displaystyle\frac{N^T_{b}I_{y}N_{b}}{N^T_{b}I_{y}N_{b}-z^T_1I_{y}z_1}\\
\notag\Xi_{\psi}=&-\left(\lambda_{\text{min}}(K_1)-\displaystyle\frac{\sigma_3}{12}\overline{N_c}^2\left(||z_s||\right)\right)N_b^TI_{\psi}N_b\\
&\text{log}\displaystyle\frac{N^T_{b}I_{\psi}N_{b}}{N^T_{b}I_{\psi}N_{b}-z^T_1I_{\psi}z_1}
\end{align}
Combining (\ref{55}),(\ref{56}), (\ref{57}) and (\ref{58}), (\ref{51}) can be revised as
\begin{align}
\hspace{-5mm}
\label{54}
\notag\dot{V}_2\leq & \Xi_{\text{x}}+\Xi_{\text{y}}+\Xi_{\psi}-[\lambda_{\text{min}}(\Gamma_1)-\displaystyle\frac{\sigma_3}{4}\overline{N_c}^2(||z_s||)]z^T_2z_2\\
\notag&-[\lambda_{\text{min}}(K_2)-\displaystyle\frac{\sigma_3}{4}\overline{N_c}^2(||z_s||)]S^TS-[\lambda_{\text{min}}(\Theta_1)\\
\notag&-\displaystyle\frac{\sigma_3}{4}\overline{N_c}^2(||z_s||)]z_f^Tz_f+\displaystyle\frac{\sigma_1\overline{M^{-1}}^2}{4}||z_2||^2\\
\notag&+\left[\displaystyle\frac{1}{\sigma_1}+\displaystyle\frac{\sigma_3}{4}\overline{N_c}^2(||z_s||)\right]||z_{\tau}||^2+\displaystyle\frac{\sigma_2\overline{(-\Gamma_1-I)}^2}{4}\\
\notag&||z_2||^2+\displaystyle\frac{1}{\sigma_2}||z_f||^2+\displaystyle\frac{1}{\sigma_3}||S||^2+\displaystyle\frac{\sigma_4}{4}\overline{d_1}^2+\displaystyle\frac{1}{\sigma_4}||S||^2\\
&+\upsilon t_d||\dot{\tau}^{'}(\theta)||^2-\upsilon\int^t_{t-t_d}||\dot{\tau}^{'}(\theta)||^2d\theta
\end{align}
Cauchy-Schwarz inequality gives the upper bound of $||z_{\tau}||$ as
\bea
\label{60}
||z_{\tau}||^2 \leq t_d \int^t_{t-t_d}||\dot{\tau}^{'}(\theta)||^2d\theta
\eea
Moreover, it can be proven that
\bea
\label{61}
\int^t_{t-t_d}\bigg[\int^t_w||\dot{\tau}^{'}(\theta)||^2d\theta\bigg]dw \leq t_d\int^t_{t-t_d}||\dot{\tau}^{'}(\theta)||^2d\theta
\eea
With (\ref{60}) and (\ref{61}), (\ref{54}) becomes
\begin{align}
\hspace{-7mm}
\label{62}
\notag\dot{V}_2\leq &\Xi_x+\Xi_y+\Xi_{\psi}-\bigg[\lambda_{\text{min}}(\Gamma_1)-\displaystyle\frac{\sigma_1\overline{M^{-1}}^2}{4}-\displaystyle\frac{\sigma_3}{4}\overline{N_c}^2(||z_s||)\\
\notag&-\displaystyle\frac{\sigma_2\overline{(-\Gamma_1-I)}^2}{4}\bigg]z^T_2z_2-\bigg[\lambda_{\text{min}}(K_2)-\displaystyle\frac{1}{\sigma_3}-\displaystyle\frac{1}{\sigma_4}-\displaystyle\frac{\sigma_3}{4}\\
\notag&\overline{N_c}^2(||z_s||)\bigg]S^TS-\bigg[\lambda_{\text{min}}(\Theta_1)-\displaystyle\frac{1}{\sigma_2}-\displaystyle\frac{\sigma_3}{4}\overline{N_c}^2(||z_s||)\bigg]\\
\notag&z_f^Tz_f+\displaystyle\frac{\sigma_4}{4}\overline{d_1}^2+\upsilon t_d ||\dot{\tau}^{'}(\theta)||^2-\bigg[\upsilon-\displaystyle\frac{t_d}{\sigma_1}\\
\notag&-\displaystyle\frac{t_d\sigma_3}{4}\overline{N_c}^2(||z_s||)\bigg]\int^t_{t-t_d}||\dot{\tau}^{'}(\theta)||^2d\theta\\
\notag\leq &\Xi_x+\Xi_y+\Xi_{\psi}-\bigg[\lambda_{\text{min}}(\Gamma_1)-\displaystyle\frac{\sigma_1\overline{M^{-1}}^2}{4}-\displaystyle\frac{\sigma_3}{4}\overline{N_c}^2(||z_s||)\\
\notag&-\displaystyle\frac{\sigma_2\overline{(-\Gamma_1-I)}^2}{4}\bigg]z^T_2z_2-\bigg[\lambda_{\text{min}}(K_2)-\displaystyle\frac{1}{\sigma_3}-\displaystyle\frac{1}{\sigma_4}\\
\notag&-\displaystyle\frac{\sigma_3}{4}\overline{N_c}^2(||z_s||)\bigg]S^TS-\bigg[\lambda_{\text{min}}(\Theta_1)-\displaystyle\frac{1}{\sigma_2}\\
\notag&-\displaystyle\frac{\sigma_3}{4}\overline{N_c}^2(||z_s||)\bigg]z_f^Tz_f-\bigg[\displaystyle\frac{\upsilon}{t_d}-\displaystyle\frac{1}{\sigma_1}-\displaystyle\frac{\sigma_3}{4}\overline{N_c}^2(||z_s||)\bigg]\\
\notag&\int^t_{t-t_d}\bigg[\int^t_w||\dot{\tau}^{'}(\theta)||^2d\theta\bigg]dw+\displaystyle\frac{\sigma_4}{4}\overline{d_1}^2+\upsilon t_d||\dot{\tau}^{'}(\theta)||^2\\
\leq & -\rho_c V_2+\beta_c
\end{align}
where $\rho_c, \beta_c>0$ and they satisfy $\rho_c=\text{min}\bigg[2\Xi_x, 2\Xi_y,
2\Xi_{psi}, 2(\lambda_{\text{min}}(\Gamma_1)-\displaystyle\frac{\sigma_1\overline{M^{-1}}^2}{4}-\displaystyle\frac{\sigma_3}{4}\overline{N_c}^2(||z_s||)-\displaystyle\frac{\sigma_2\overline{(-\Gamma_1-I)}^2}{4}), 2(\lambda_{\text{min}}(K_2)-\displaystyle\frac{1}{\sigma_3}-\displaystyle\frac{1}{\sigma_4}-\displaystyle\frac{\sigma_3}{4}\overline{N_c}^2(||z_s||))/\lambda_{\text{max}}(M), 2(\lambda_{\text{min}}(\Theta_1)-\displaystyle\frac{1}{\sigma_2}-\displaystyle\frac{\sigma_3}{4}\overline{N_c}^2(||z_s||)), (\displaystyle\frac{1}{t_d}-\displaystyle\frac{1}{\sigma_1\upsilon}-\displaystyle\frac{\sigma_3}{4\upsilon}\overline{N_c}^2(||z_s||))\bigg]$
with the tuning parameters are selected $\lambda_{\text{min}}(K_1)>\displaystyle\frac{\sigma_3}{12}\overline{N_c}^2(||z_s||)$, $\lambda_{\text{min}}(\Gamma_1)+\displaystyle\frac{\sigma_2\overline{(-\Gamma_1-I)}^2}{4}>\displaystyle\frac{\sigma_1\overline{M^{-1}}^2}{4}+\displaystyle\frac{\sigma_3}{4}\overline{N_c}^2(||z_s||)$, $\lambda_{\text{min}}(K_2)>\displaystyle\frac{1}{\sigma_3}+\displaystyle\frac{1}{\sigma_4}+\displaystyle\frac{\sigma_3}{4}\overline{N_c}^2(||z_s||)$, $\lambda_{\text{min}}(\Theta_1)>\displaystyle\frac{1}{\sigma_2}+\displaystyle\frac{\sigma_3}{4}\overline{N_c}^2(||z_s||)$, $\displaystyle\frac{\upsilon}{t_d}>\displaystyle\frac{1}{\sigma_1}+\displaystyle\frac{\sigma_3}{4}\overline{N_c}^2(||z_s||)$.
$\beta_c=\displaystyle\frac{\sigma_4}{4}\overline{d_1}^2+\upsilon t_d||\dot{\tau}^{'}(\theta)||^2$.

\begin{lemma} \label{lemm2} \cite{ge2002direct} \cite{tee2006control} For bounded initial conditions, if there exists a $C^1$ continuous and positive definite Lyapunov function $V(x)$ satisfying $v_1(||x||)\leq V(x) \leq v_2(||x||)$, such that $\dot{V} \leq -\alpha V(x)+\beta$, where $v_1$, $v_2$: $\text{R}^n \rightarrow \text{R}$ are class K functions and $\alpha, \beta>0$, then the solution $x(t)$ is uniformly bounded.
\end{lemma}
\begin{remark}
Combining Lemma \ref{lemm2}, Remark \ref{rem5} and (\ref{37})-(\ref{62}), the Semi-Globally Uniform Boundness (SGUB) of all the signals are guaranteed under the existence of input delay. In addition, the tracking error is regulated as $|z_1|\leq N_b$.
\end{remark}
\subsection{Design of control after alarm}
For robust control under large wave-induced force, we consider the model in (\ref{overallmodel1}) (\ref{overallmodel2}). Similar to the control before alarm, the wind force is estimated for the feedforward control. Thus, the dynamic model of (\ref{overallmodel2}) can be rewritten as
\be
M\dot{\nu}+C(\nu)\nu+D(\nu)\nu+g(\eta)=\tau^{'}(t-t_d)+\gamma(t-T)\tau_{\text{wave}}+d_1
\ee
For simplicity, in the following proof, the term $\gamma(t-T)\tau_{\text{wave}}$ will be replaced by $\tau_{\text{wave}}$.
The first step of the control design after alarm is the same with \emph{step 1}. And all the proof before (\ref{42}) remain the same, (\ref{44}) will be changed into
\begin{align}
\hspace{-5mm}
\notag M\dot{S}=&M\dot{\alpha}_c+M_s-\tau_{\text{wave}}-d_1+N_c-\tau^{'}(t)\\
&-K_2S-K_2z_2-(S^{T})^{+}\dot{S}^Tz_2
\end{align}
To estimate the unknown wave force, a RBF neural network is applied.
\bea
\tau_{\text{wave}}=W^{*T}_{c}S_c(Z_c)+\epsilon_c
\eea
Denote $\hat{W}_c, W_c^{*}, \epsilon_c$ as the estimated weights, optimal weights and approximation error respectively. $Z_c$ is the input vector to the neural network. The details about $Z_c$ will be introduced in the simulation section. Design the update law of the NN weights to be
\bea
\label{weightcon}
\dot{\hat{W}}_{ci}=-\Upsilon_i(S_{ci}(Z_c)S_i+\xi_i\hat{W}_{ci})
\eea
Control input under this condition should be augmented into
\bea
\label{tau_mc}
\tau^{'}_m(t)=\tau^{'}(t)-\hat{W}^{T}_{c}S_c(Z_c)
\eea
The control law in (\ref{tau_mc}) is able to guarantee the SGUB of all the close-loop system states. \\
\emph{Proof} The proof is very trivial and similar to that in ``control before alarm" section, thus, ignore here.
\section{Optimal Thrust Allocation for Dynamic Positioning}
\subsection{Problem Formulation for Thrust Allocation}
\begin{figure}[htb]
  \centering
  \includegraphics[width=90mm,height=33mm]{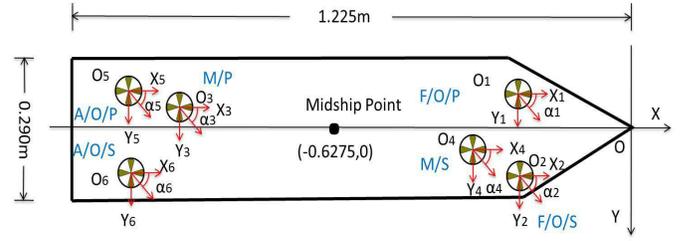}
  \caption{Thruster layout and coordinate system}
   \label{AV2}
\end{figure}

\begin{figure}[htb]
  \centering
  \includegraphics[width=90mm,height=35mm]{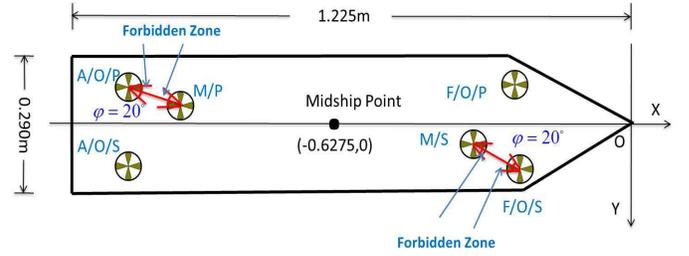}
  \caption{Definition of The Forbidden Zone}  \label{Forbid1}
\end{figure}

This section will give an optimal solution in terms of individual thruster to achieve required resultant force along axis $X$ and $Y$ and resultant torque $M_z$. The AV DP system is compounded by 6 nozzle thrusters. Each of them can rotate the full $360^{\circ}$ to generate thrust in any direction. The six thrusters are grouped in pairs and their layout are presented in Fig. \ref{AV2}.\\
\indent In addition, to avoid thruster-thruster interaction, a forbidden zone \cite{wei2013quadratic} of $20 ^{\circ}$ is considered to increase the propelling efficiency. The forbidden zone in this paper is depicked as Fig. \ref{Forbid1}.

The resulting force and moment generated by the 6 thrusters in surge, sway and yaw direction are given by
\begin{equation}
F_x=\sum^6_{i=1}\cos\alpha_iu_i=A_{f_x}(\alpha)u,
F_y=\sum^6_{i=1}\sin\alpha_iu_i=A_{f_y}(\alpha)u
\end{equation}
\begin{align}
\hspace{-5mm}
\notag M_z=&(l_{1x}\cos\alpha_1+l_{1y}\sin\alpha_1)u_1+(-l_{2x}\cos\alpha_2+l_{2y}\sin\alpha_2)\\
\notag&u_2+(l_{3x}\cos\alpha_3-l_{3y}\sin\alpha_3)u_3+(-l_{4x}\cos\alpha_4\\
\notag&+l_{4y}\sin\alpha_4)u_4+(l_{5x}\cos\alpha_5-l_{5y}\sin\alpha_5)u_5+(-l_{6x}\\
&\cos\alpha_6-l_{6y}\sin\alpha_6)u_6=A_{M}(\alpha)u
\end{align}
where $l_{\text{ix}}$ and $l_{\text{iy}}$ (i=1,2,...,6) are the moment arm along X and Y direction of the $i$th thruster. $\alpha_i$ and $u_i$ are the rotation angle and the magnitude of thrust produced by the $i$ th thruster. $\alpha_i$s and $u_i$s are merged as $\alpha=[\alpha_1,\alpha_2,\alpha_3,\alpha_4,\alpha_5,\alpha_6]^T$ and $u=[u_1,u_2,u_3,u_4,u_5,u_6]^T$. The sum of generalized propelling forces on the vessel from the thrusters are modelled as
\bea
\tau=\mathcal{T}(\alpha)u
\eea
where $\mathcal{T}(\alpha)=[A_{f_x}(\alpha), A_{f_y}(\alpha), A_{M}(\alpha)]^T$. $\tau$ is the command signal which is the combination of feedforward wind force compensation and the feedback control effort designed in the last section. The cost function is formulated as
\begin{equation}
J=\text{min}\{u^T\mathcal{Q}u+(\alpha-\alpha_0)^T\mathcal{P}(\alpha-\alpha_0)+o^T\mathcal{R}o\}
\end{equation}
subject to:
\bea
\label{ec}
\mathcal{T}(\alpha)u=\tau+o, \quad
\underline{u}\leq u \leq \overline{u}, \quad \underline{\alpha} \leq \alpha \leq \overline{\alpha}
\eea
\bea
\underline{\Delta \alpha}\leq \alpha-\alpha_0 \leq \overline{\Delta\alpha}
\eea
where $u^T\mathcal{Q}u$ represents power consumption and $\mathcal{Q} \in \mathbb{R}^{6\times6}$ is a positive weight matrix. The second term of the cost function is used to guarantee a minimum rotation angle of each thruster in a single sampling interval with positive weights $\mathcal{P} \in \mathbb{R}^{6\times6}$.
$\alpha_0 \in \mathbb{R}^{6\times1}$ represent current rotated angle of the thrusters. $o^T\mathcal{R}o$ penalizes the error $o \in \mathbb{R}^{3\times1}$ between the commanded and achieved generalized force. The weight $\mathcal{R} \in \mathbb{R}^{3\times3}$ should be chosen sufficiently large so that the error is necessarily small.
  $\underline{u}\leq u \leq \overline{u}$ denotes the limit of thrust in this case. $\underline{\alpha} \leq \alpha \leq \overline{\alpha}$ restricts the feasible working zone, in this case, $20^{\circ}$ forbidden zone is considered.
$\underline{\Delta \alpha}\leq \alpha-\alpha_0 \leq \overline{\Delta\alpha}$ gives the constrain of azimuth speed.
\subsection{Locally Convex Reformulation}
The above formulation usually contributes to a nonlinear non-convex problem which requires large computations to search the solution. The main reason is the nonlineariy of the equality constraint (\ref{ec}). To simplify the solution search process, a locally convex quadratic programming reformulation is suggested.
Since in dynamic positioning the azimuth angles are required to be slowly varying near the position in last sampling time instant $\alpha_0$ and similar situation holds for the output thrust, linearization of the equality constraint at the current thruster state (output thrust and angle) is reasonable. Therefore, the optimization problem can be reformulated as follows.
\be
\label{ncf}
\begin{split}
J=&\text{min}\{(u_0+\Delta u)^T\mathcal{Q}(u_0+\Delta u)+\Delta\alpha^T\mathcal{P}\Delta\alpha+o^T\mathcal{R}o\}\\
=&\text{min}\{\Delta u^T \mathcal{Q} \Delta u+\Delta\alpha^T\mathcal{P}\Delta\alpha+o^T\mathcal{R}o+(2\mathcal{Q}^Tu_0)^T\Delta u\}
\end{split}
\ee
subject to:
\bea
\mathcal{T}(\alpha_0)\Delta u+\left.\displaystyle\frac{\partial}{\partial \alpha}(\mathcal{T}(\alpha)u)\right|_{\alpha_0, u_0 }\Delta \alpha-o=\tau-\mathcal{T}(\alpha_0)u_0
\eea
\bea
\label{optcon1}
\underline{u}-u_0\leq \Delta u \leq \overline{u}-u_0, \quad
\underline{\alpha}-\alpha_0\leq \Delta \alpha \leq \overline{\alpha}-\alpha_0
\eea
\bea
\label{optcon2}
\underline{\Delta \alpha}\leq \Delta \alpha \leq \overline{\Delta\alpha}
\eea
The optimization problem above can be rewritten as the following more compact form.
\bea
\label{nncf}
J=\text{min}\{\displaystyle\frac{1}{2}\mathcal{U}^T\mathcal{K}\mathcal{U}+\mathcal{W}^T\mathcal{U}\}
\eea
s.t.
\bea
\label{nnc}
\mathcal{M}\mathcal{U}=\mathcal{Y},\quad
\underline{\mathcal{U}}\leq \mathcal{U}\leq \overline{\mathcal{U}}
\eea
where
$\mathcal{U}=\begin{bmatrix}\Delta u^T, \Delta \alpha^T, o^T\end{bmatrix}^T \in \Omega_{\mathcal{U}}$, $\Omega_{\mathcal{U}}:=\big\{\mathcal{U}\in \mathbb{R}^{6\times1}|\underline{\mathcal{U}}\leq \mathcal{U} \leq \overline{\mathcal{U}}\big\}$. Other vectors and matrices are defined as $\mathcal{K}=\text{diag}[2\mathcal{Q}, 2\mathcal{P}, 2\mathcal{R}]$, $\mathcal{W}=\begin{bmatrix}(2\mathcal{Q}^Tu_0)^T, O_{1\times6}, O_{1\times3}\end{bmatrix}^T$, \\
$\mathcal{M}=\begin{bmatrix}\mathcal{T}(\alpha_0), \left.\displaystyle\frac{\partial}{\partial \alpha}(\mathcal{T}(\alpha)u)\right|_{\alpha_0, u_0}, -I\end{bmatrix}$, $\mathcal{Y}=\tau-\mathcal{T}(\alpha_0)u_0$,\\
$\underline{\mathcal{U}}=\begin{bmatrix}(\underline{u}-u_0)^T, \text{max}((\underline{\alpha}-\alpha_0),\underline{\Delta \alpha})^T, \underline{o}^T\end{bmatrix}^T$, $\overline{\mathcal{U}}=\begin{bmatrix}(\overline{u}-u_0)^T, \text{min}((\overline{\alpha}-\alpha_0), \overline{\Delta \alpha})^T, \overline{o}^T\end{bmatrix}^T$.
\subsection{LVIPDNN Optimization}
To solve online the linear Quadratic Program (QP) problem shown in (\ref{nncf})-(\ref{nnc}), a simplified gradient LVIPDNN is adopted. Firstly, the above optimization problem is converted to the lagrangian dual problem. Follow \cite{bazaraa2013nonlinear}, the dual problem is to maximize $\mathcal{H}(\mathcal{U})$ with
\begin{align}
\hspace{-5mm}
\notag \mathcal{H}(\mathcal{U})=&\text{inf}\{\displaystyle\frac{1}{2}\mathcal{U}^T\mathcal{K}\mathcal{U}+\mathcal{W}^T\mathcal{U}+\mathcal{V}^T(\mathcal{Y}-\mathcal{M}\mathcal{U})\\
&+\underline{\mathcal{L}}^T(\underline{\mathcal{U}}-\mathcal{U})+\overline{\mathcal{L}}^T(\mathcal{U}-\overline{\mathcal{U}})\}
\end{align}
where $\mathcal{V}\in \Omega_{\mathcal{V}}$, $\Omega_{\mathcal{V}}:=\big\{\mathcal{V}\in \mathbb{R}^{3\times1}|-\overline{\mathcal{V}}\leq \mathcal{V} \leq \overline{\mathcal{V}}\big\}$. $\overline{\mathcal{V}}$ is a sufficiently large constant vector to represent $+\infty$. $\underline{\mathcal{L}}$ and $\overline{\mathcal{L}} \in \mathbb{R}^{6\times1}$ are dual-decision variables. The necessary and sufficient condition for a minimum is the vanishing of the gradient
\be
\label{93}
\displaystyle\frac{ \partial \mathcal{H}(\mathcal{U})}{\partial \mathcal{U}}=\mathcal{K}\mathcal{U}+\mathcal{W}-\mathcal{M}^T\mathcal{V}-\underline{\mathcal{L}}+\overline{\mathcal{L}}=0
\ee
With this condition, we can further obtain the following equation.
\be
-\mathcal{U}^T\mathcal{K}\mathcal{U}=\mathcal{W}^T\mathcal{U}-\mathcal{V}^T\mathcal{M}\mathcal{U}-\underline{\mathcal{L}}^T\mathcal{U}+\overline{\mathcal{L}}^T\mathcal{U}
\ee
The dual quadratic formulation can be derived
\bea
J_d=\text{max}\{-\displaystyle\frac{1}{2}\mathcal{U}^T\mathcal{K}\mathcal{U}+\mathcal{V}^T\mathcal{Y}+\underline{\mathcal{L}}^T\underline{\mathcal{U}}-\overline{\mathcal{L}}^T\overline{\mathcal{U}}\}
\eea
s.t. (\ref{93}) with $\mathcal{V}$, $\underline{\mathcal{L}}$, $\overline{\mathcal{L}}$ $\geq 0$. Our objective is to convert the QP problem into a set of LVIs by finding a primal-dual equilibrium vector $\mathcal{U}^{*}\in \Omega_{\mathcal{U}}$, $\mathcal{V}\in \Omega_{\mathcal{V}}$ \cite{zhang2004unified},
such that
\bea
\label{96}
(\mathcal{U}-\mathcal{U}^{*})^T(\mathcal{K}\mathcal{U}^{*}+\mathcal{W}-\mathcal{M}^T\mathcal{V}^{*})\geq0
\eea
Similarly, the LVIs for (\ref{nnc}) is
\bea
\label{97}
(\mathcal{V}-\mathcal{V}^{*})^T(\mathcal{M}\mathcal{U}^*-\mathcal{Y})\geq 0
\eea
Combining (\ref{96}) and (\ref{97}), the LVIs for the whole system can be rewritten as
\begin{align}
\hspace{-5mm}
\notag&\left(\begin{bmatrix}\mathcal{U}\\\mathcal{V}\end{bmatrix}-\begin{bmatrix}\mathcal{U}^{*}\\\mathcal{V}^{*}\end{bmatrix}\right)^T\left(\begin{bmatrix}\mathcal{K}&-\mathcal{M}^T\\\mathcal{M}&0\end{bmatrix}\begin{bmatrix}\mathcal{U}^{*}\\\mathcal{V}^{*}\end{bmatrix}+\begin{bmatrix}\mathcal{W}\\-\mathcal{Y}\end{bmatrix}\right)\\
&=\left(\mathcal{Z}-\mathcal{Z}^{*}\right)^T\left(\mathcal{E}\mathcal{Z}^{*}+\mathcal{S}\right)\geq0
\end{align}
where $\mathcal{Z}=\begin{bmatrix}\mathcal{U}\\\mathcal{V}\end{bmatrix}\in \Omega_{\mathcal{Z}}=\Omega_{\mathcal{U}}\times\Omega_{\mathcal{V}}$, $\mathcal{E}=\begin{bmatrix}\mathcal{K}&-\mathcal{M}^T\\\mathcal{M}&0\end{bmatrix}$ and $\mathcal{S}=\begin{bmatrix}\mathcal{W}\\-\mathcal{Y}\end{bmatrix}$.
The following piecewise linear equation is applied to reformulate the above LVIs \cite{zhang2002dual}.
\bea
\label{99}
\mathcal{G}_{\Omega_{\mathcal{Z}}}\left(\mathcal{Z}-\left(\mathcal{E}\mathcal{Z}+\mathcal{S}\right)\right)-\mathcal{Z}=0
\eea
where $\mathcal{G}_{\Omega_{\mathcal{Z}}}(\bullet)$ denotes the projection operator on $\Omega_{\mathcal{Z}}$ with the following definition.
\bea
\mathcal{G}_{\Omega_{\mathcal{Z}}}(\mathcal{B})\begin{cases}\underline{\mathcal{B}}, \qquad \text{if} \qquad \mathcal{B} < \underline{\mathcal{B}}\\
\mathcal{B}, \qquad \text{if} \qquad \underline{\mathcal{B}}\leq \mathcal{B}\leq \overline{\mathcal{B}}\\
\overline{\mathcal{B}}, \qquad \text{if} \qquad \mathcal{B} > \overline{\mathcal{B}}
\end{cases}
\eea
The following dynamical system is developed for (\ref{99}) according to dynamic-solver design approach \cite{zhang2004unified} \cite{zhang2008simplified}.
\bea
\label{optiZ}
\dot{\mathcal{Z}}=\Gamma_{\mathcal{Z}}\left(I+\mathcal{E}^T\right)\{\mathcal{G}_{\Omega_{\mathcal{Z}}}\left(\mathcal{Z}-\left(\mathcal{E}\mathcal{Z}+\mathcal{S}\right)\right)-\mathcal{Z}\}
\eea
$\Gamma_{\mathcal{Z}} \in \mathbb{R}^{18\times18}$ is positive parameter used to tune the convergence rate \cite{li2014contact}.
\begin{theorem}
Assume the existence of optimal solution to the locally convex QP problem in (\ref{nncf})-(\ref{nnc}). The output of the search law (\ref{optiZ}) is globally exponentially convergent to the optimal solution $\mathcal{U}^{*}$.
\end{theorem}
\section{Simulation Study}
In this section, a supply vessel replica-Cybership \uppercase\expandafter{\romannumeral2} in the marine control laboratory of Norwegian University of Science and Technology (NTNU) \cite{skjetne2005adaptive} is considered as the case study to evaluate the performance of the proposed control scheme.
\subsection{Environmental Forces}
\subsubsection{Wind Forces}
The wind force model is as presented in (\ref{windmod}). The wind direction is along $X_0$ with the velocity of 16m/s. The peak of wind drag coefficients are selected as $[C_x,C_y,C_N]^T=[0.1,0.14,0.1]^T$.
\subsubsection{Wave Forces}
In this section, the wave forces indicate the wave-induced drift forces. These forces refer to the nonzero slowly varying components of the total wave-induced force. In this paper, we assume that the high-frequency components,i.e., the first-order wave-induced forces are filtered out by filters in advance and in DP system, no control is applied to handle the high-frequency motion. The model of wave drift forces are considered as follow \cite{fossen2011handbook}.
\be
\label{wavedrift}
\tau^{\left[dof\right]}_{\text{wave}}=\sum^N_{k=1} \rho_{\text{water}}g_{v}\left|F_{\text{wave2}}(\omega_k,\beta_r)\right|A_k^2\text{cos}\big(\omega_{e}(U,\omega_k,\beta_r)t+\epsilon_k\big)
\ee
where, $\left|F_{\text{wave2}}(\omega_k,\beta_r)\right|$ is the amplitude of the mean drift force. $\omega_k$ and $\beta_r$ are wave frequencies and the angle between the heading of the vessel and the attack direction of the wave. The wave comes from the same direction with the wind, i.e, $\beta_{\text{wave}}=0$. The calculation of $\left|F_{\text{wave2}}(\omega_k,\beta_r)\right|$ should be obtained by complex RAO analysis. For simplicity, we adopt a sinusoidal function to estimate it. $A_k$ satisfies $\displaystyle\frac{1}{2}A_k^2=S(\omega_k)\Delta\omega$. $S(\omega)$ is the JONSWAP wave spectrum. The dominant wave frequency is denoted as $\omega_o$ and $\omega_o=6\times10^{-4}\text{rad/s}$. The encounter frequency $\omega_{e}$is defined as $\omega_{e}(U,\omega_{o},\beta)=\left|\omega_{o}-\displaystyle\frac{\omega^2_{o}}{g_v}U\text{cos}(\beta)\right|$. $U$ is the total speed of the ship. $\epsilon_k$ is the random phase angle chosen within the range of $\left[-0.2,0.2\right] \text{rad/s}$.\\
\indent In this simulation, we assume that during the beginning 10s, the sea is calm and the state becomes moderate at 10s, which triggers the rotation motion of the FPSO. While, because of the shielding effect, the large wave force starts to attack the AV at 150s. After that, the drift force increases gradually and the model (\ref{wavedrift}) is activated to generate the force and moment.
\subsection{Control System Simulation Study}
In response to wind and wave force acting on the FPSO, the trajectory of FPSO is approximately a quarter round with the amplitude of 17m and frequency of 0.005rad/s. Thus, the desired trajectory of the accommodation vessel can be expressed as
\be
\begin{cases}\eta_{\text{xd}}(t)=17\text{sin}\big(0.005(t-t_{m})\big)\\
\eta_{\text{yd}}(t)=-17\text{sin}\left(0.005(t-t_{m})+\displaystyle\frac{\pi}{2}\right)\\
\eta_{\text{zd}}(t)=\displaystyle\frac{\pi}{2}-\text{arctan}\left(\displaystyle\frac{\left|\eta_{xd}\right|}{\left|\eta_{yd}\right|}\right)
\end{cases}
\ee
where $t_{m}=10s$ is the moment when the sea state changes. The initial position and velocity of the vessel are $\eta_0=\left[0, -17, \displaystyle\frac{\pi}{2}\right]^T$ and $\nu_0=\left[0, 0, 0\right]^T$. The total simulation time is 324s.
\subsubsection{Sea Observer}
Initially, (\ref{besever}) is applied to approximate the position and velocity of the vessel as well as the wind force and moment before alarm. The parameters are designed as $L=5I_{6\times6}$, $C=I_{6\times6}$. $\Gamma$ and $P$ in adaptive law (\ref{adawind}) are selected as $\Gamma=\text{diag}\left\{100,600,100 \right\}$ and $P=5I_{6\times6}$ respectively. The initial condition of the observer and the wind drag coefficient estimator are designed as $X_{0}=\left[ \eta_0, \nu_0\right]^T$ and $\hat{\Phi}_0=\left[0.024, 0.056, 0.033\right]^T$.\\
\indent Due to the effect of wave-induced force, when the vessel moves out of the shadow of the FPSO, the wind force estimator would conduct overcompensation. The overcompensation provide us with adequate hint to decide when the NN compensator is on. If the mean value of the estimated wind drag coefficients in past 5 successive seconds is above 0.2, a judgement can be made that severe wave force is attacking the vessel and the NN compensation needs to be activated both in the sea observer and in the controller.\\
\indent After the compensation is triggered, since the over-compensated wind estimator cannot approximate the wave-induced forces perfectly, the update law with NN estimator (\ref{sevdet}) is applied. The network in this observer has $2^5$ nodes. The inputs $Z_{ow}$ contain $A_o, \omega_o, \beta_\text{wave}, \hat{\dot{\eta}}_x, \hat{\dot{\eta}}_y, \hat{\eta}_\phi$. Where $A_o$ denotes $A_k$ in (\ref{wavedrift}) at the point of dominant frequency $\omega_o$. The corresponding center are distributed in $\left[-0.5, 0.5\right], \left[-0.5, 0.5\right], \left[-0.5, 0.5\right], \left[-0.5, 0.5\right], \left[-0.5, 0.5\right]$ and $\left[-2, 2\right]$ respectively. The initial values of the weights are $\hat{W}_{di}=O_{2^5\times1}, \left(i=1,2,3\right)$. The updating rate in adaptive law (\ref{nndet}) is $\omega_i=0.002,\left(i=1,2,3\right)$.
\subsubsection{Robust Control}
\begin{figure}[thpb]
  \centering
\subfigure[Wind forces and moment]{\label{windformom}\includegraphics[width=0.47\hsize]{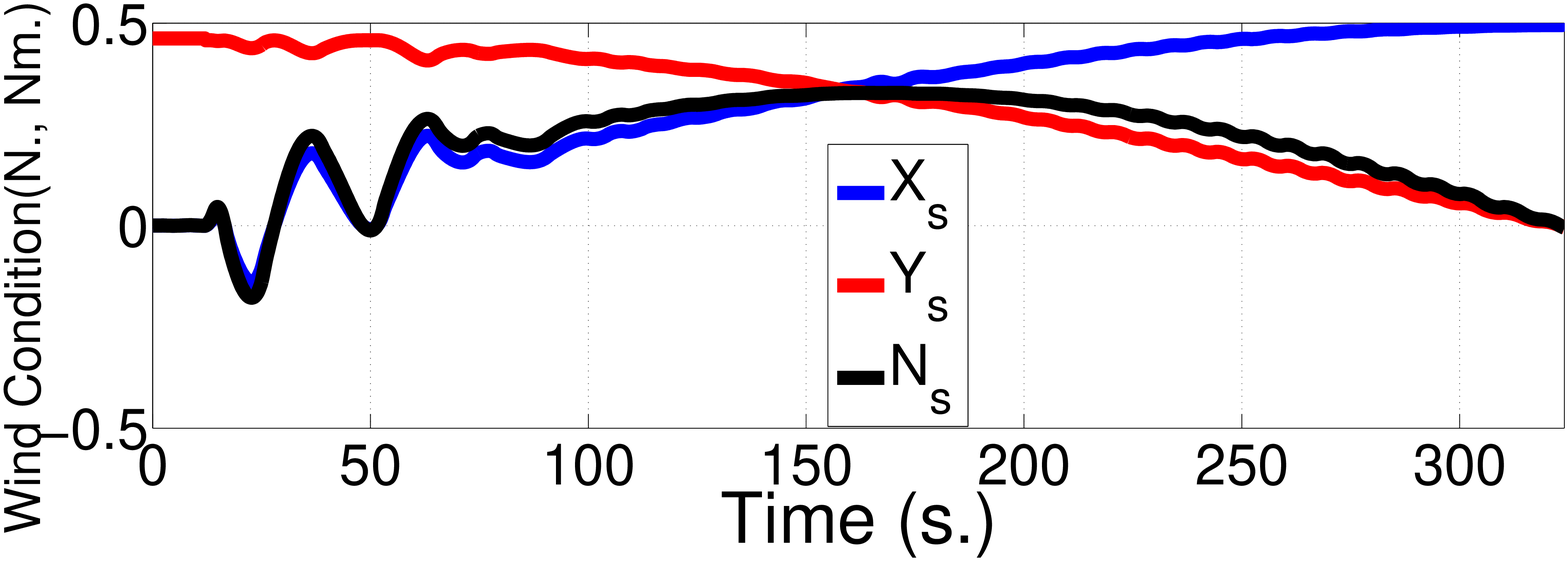}}
\subfigure[Wave forces and moment]{\label{wandformom}\includegraphics[width=0.47\hsize]{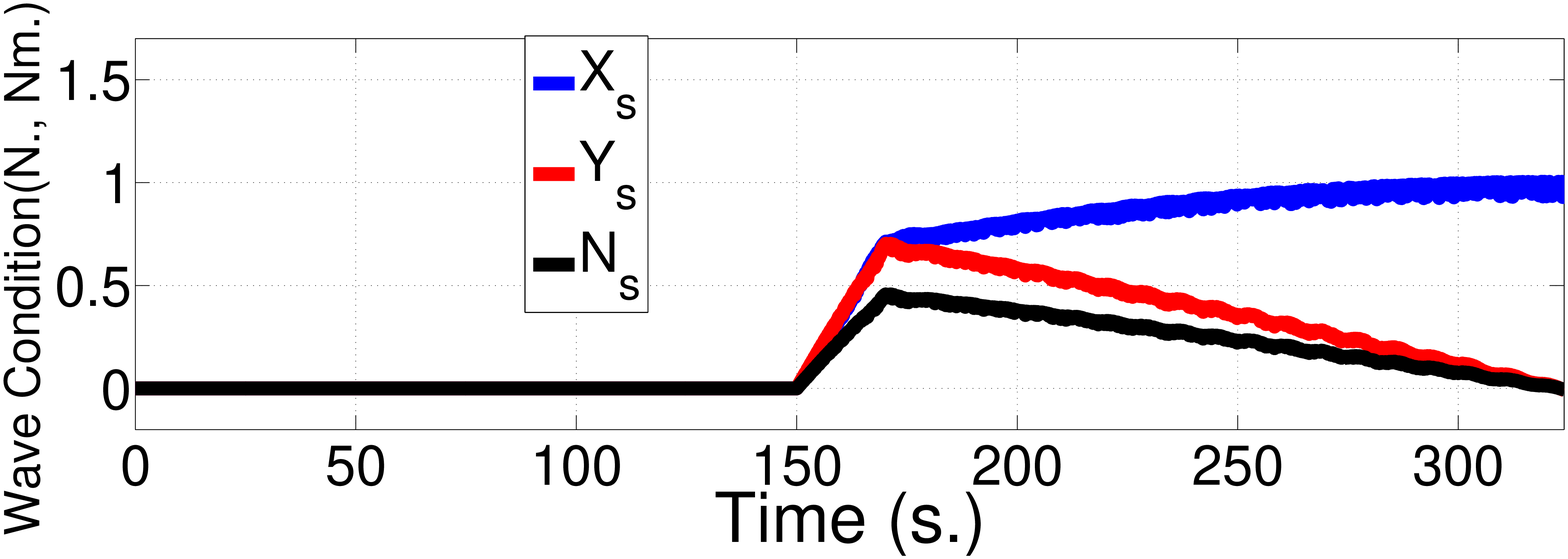}}
  \caption{Environmental forces and moment.} \label{olprate}
\end{figure}

\begin{figure}[thpb]
  \centering
\subfigure[The desired and vessel motion in surge]{\label{simetax}\includegraphics[width=0.47\hsize]{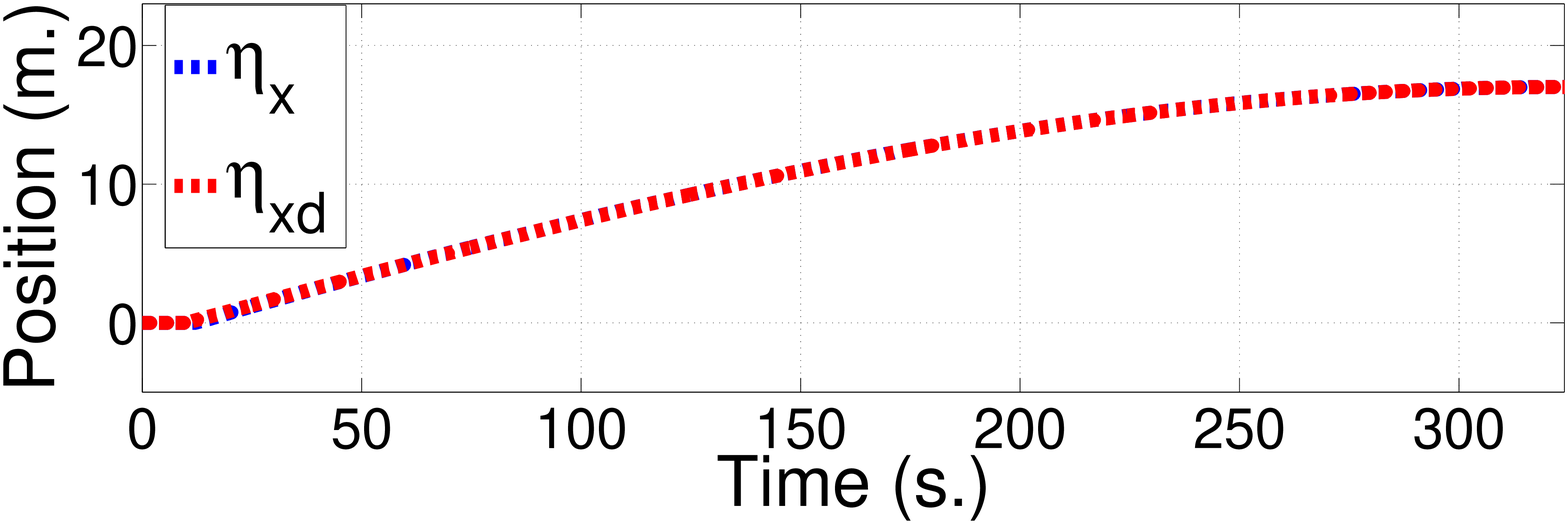}}
\subfigure[The desired and vessel motion in sway]{\label{simetay}\includegraphics[width=0.47\hsize]{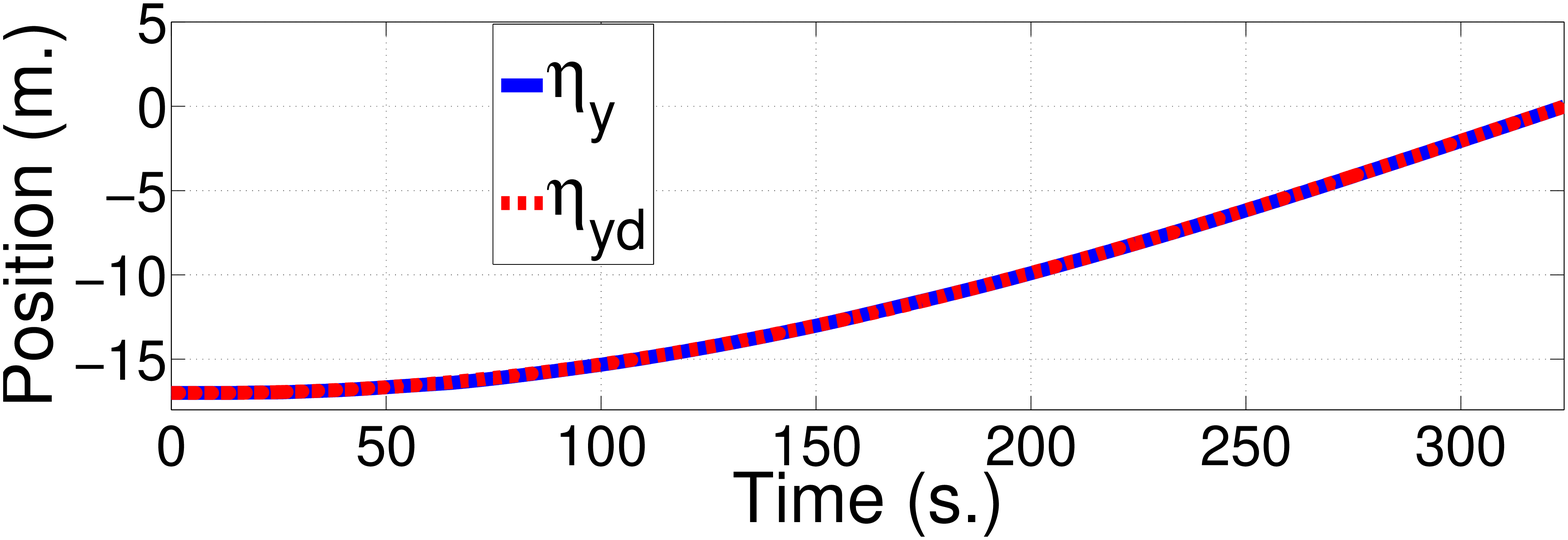}}
\subfigure[The desired and vessel motion in yaw]{\label{simetapsi}\includegraphics[width=0.47\hsize]{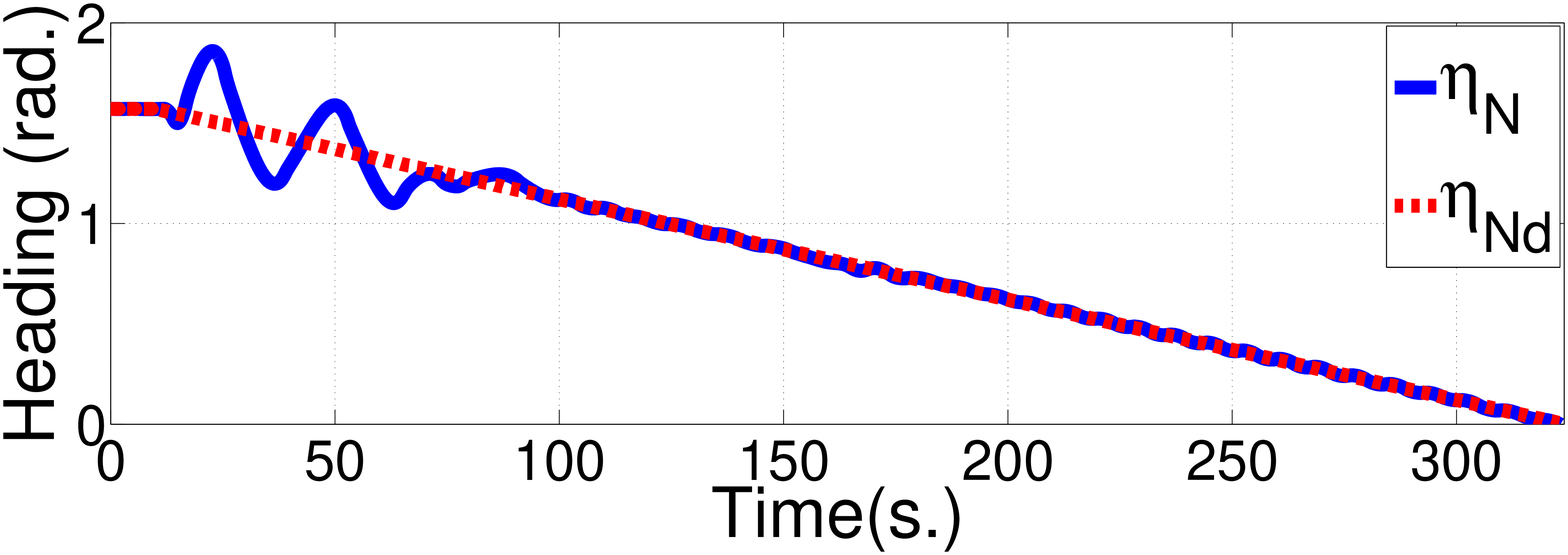}}
  \caption{Control performance in 3 DOFs.} \label{olprate11}
\end{figure}

\begin{figure}[thpb]
  \centering
\subfigure[Tracking errors in surge and sway]{\label{simerrxy}\includegraphics[width=0.47\hsize]{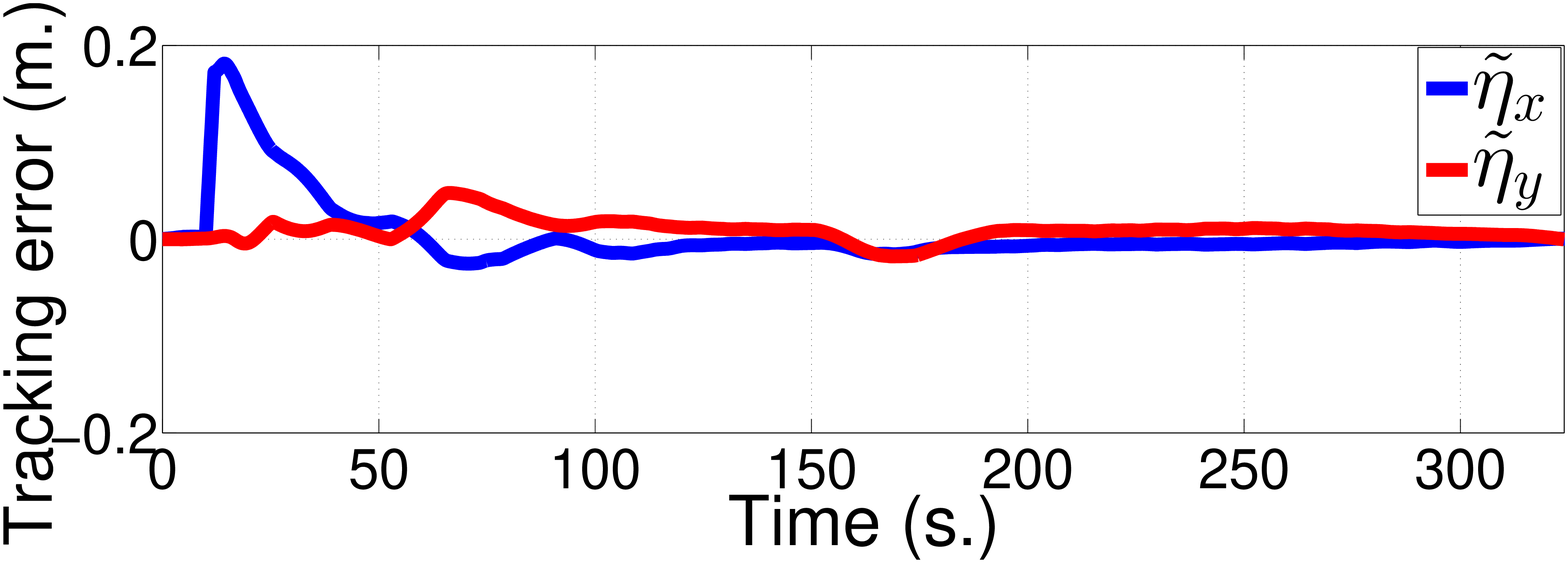}}
\subfigure[Tracking errors in yaw]{\label{simerrpsi}\includegraphics[width=0.47\hsize]{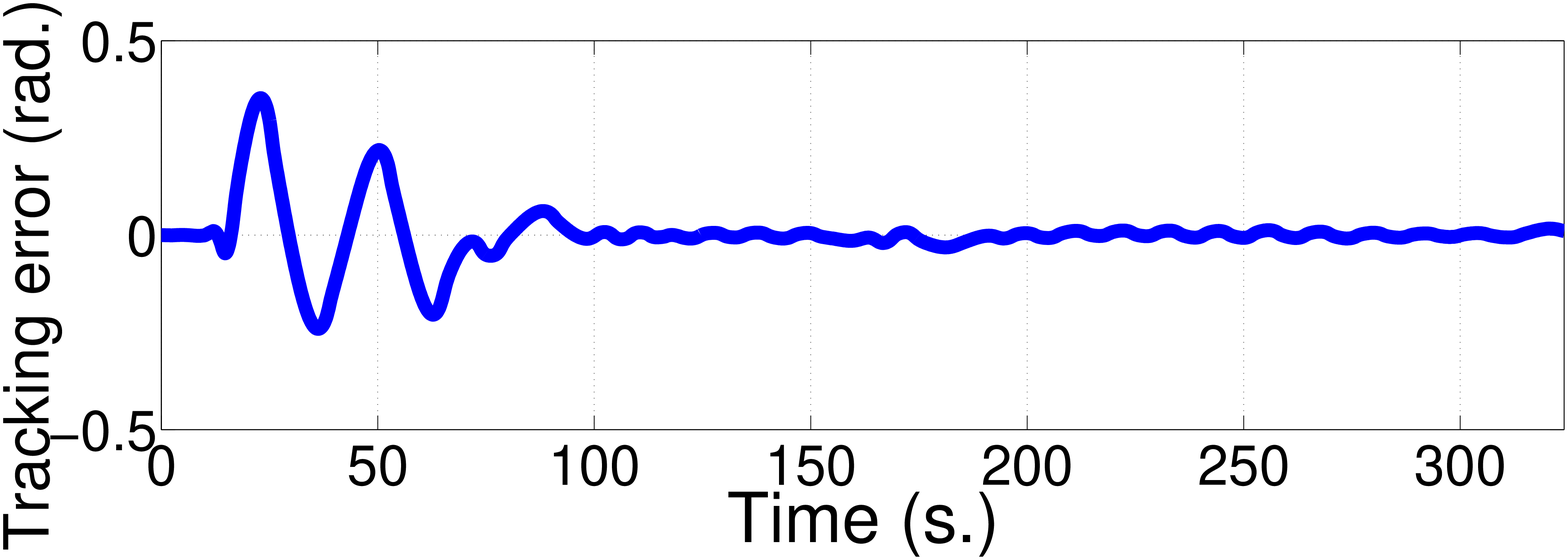}}
  \caption{Tracking errors.} \label{olprate111}
\end{figure}

\begin{figure}[htb]
  \centering
  \includegraphics[width=90mm,height=35mm]{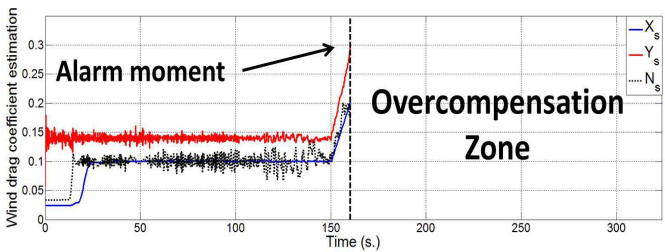}
  \caption{Estimated wind drag coefficient} \label{wce}
\end{figure}

\begin{figure}[htb]
  \centering
  \includegraphics[width=90mm,height=35mm]{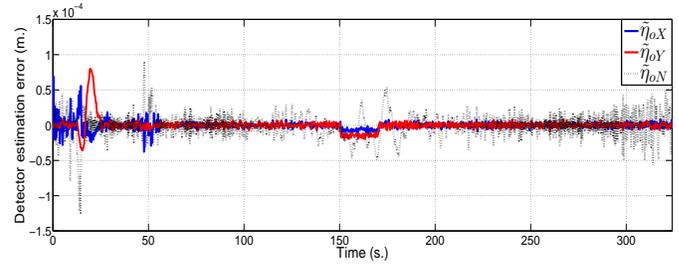}
  \caption{State estimation error of the sea observer} \label{simdesterr}
\end{figure}

Before the switching command is received from sea observer, dynamic model in (\ref{modelconbe}) is considered. The input time delay $t_d$ is 2s. The disturbance $d$ is chosen randomly between -0.05-0.05. The gangway is able to rotate $360^{\circ}$ freely, thus the tracking error constraint on yaw motion is relatively loose. $N_b$ is set to be $N_b=\left[0.3, 0.3, \displaystyle\frac{\pi}{6}\right]^T$. Control law in (\ref{tau11}) is applied with the parameters tuned as $K_1=0.001\text{diag}\left\{6, 6, 4\right\}$, $K_2=0.001\text{diag}\left\{6, 6, 4\right\}$, $\Gamma_1=0.001\text{diag}\left\{1, 1, 2\right\}$ and $\Theta=0.001\text{diag}\left\{1, 1, 1\right\}$. The initial condition of the auxiliary state is $z_{f0}=\left[0, 0, 0\right]^T$.\\
\indent When NN is required for wave force compensation, control law (\ref{tau_mc}) is activated. The network also contains $2^5$ nodes with the the center evenly distributed in $\left[-0.5, 0.5\right], \left[-0.5, 0.5\right], $ $\left[-0.5, 0.5\right], \left[-0.5, 0.5\right], \left[-0.5, 0.5\right]$ and $\left[-2, 2\right]$ respectively. The initial value of the weights are $\hat{W}_{ci}=O_{2^5\times1}, \left(i=1,2,3\right)$. The input of the network $Z_c$ include $A_o, \omega_o, \beta_\text{wave}, \dot{\eta}_x, \dot{\eta}_y, \eta_\phi$. The updating rates in (\ref{weightcon}) are tuned as $\Upsilon_i=2.2$ and $\xi_i=2.2, \left(i=1,2,3\right)$. The wind and wave forces and moment acting on the vessel can be found in Fig. (\ref{olprate}). The control performance can be seen from Fig. (\ref{olprate11})-(\ref{simdesterr}).

\subsubsection{Thrust Allocation}
\begin{figure}[htb]
  \centering
  \includegraphics[width=60mm,height=23mm]{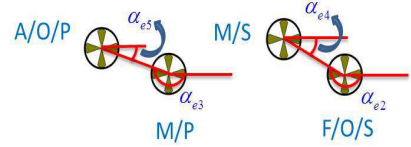}
  \caption{Definition of encounter angle} \label{encoan}
\end{figure}

\begin{figure}[thpb]
  \centering
\subfigure[Output of thrust allocation in surge]{\label{allx}\includegraphics[width=0.47\hsize]{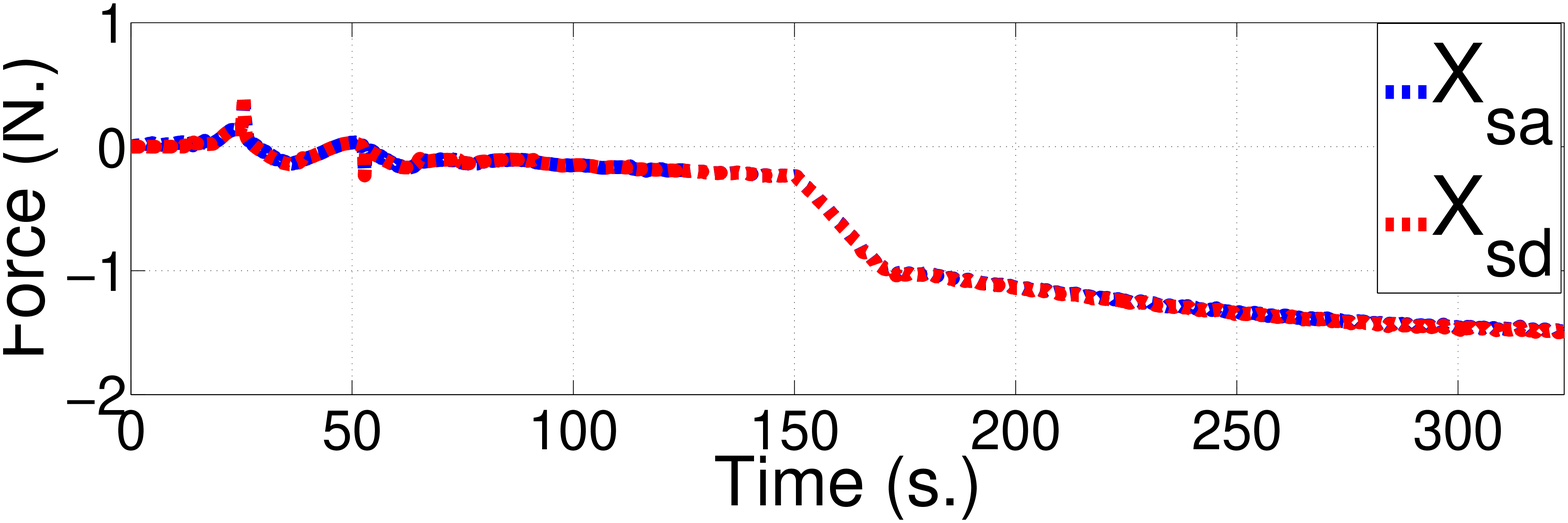}}
\subfigure[Output of thrust allocation in sway]{\label{ally}\includegraphics[width=0.47\hsize]{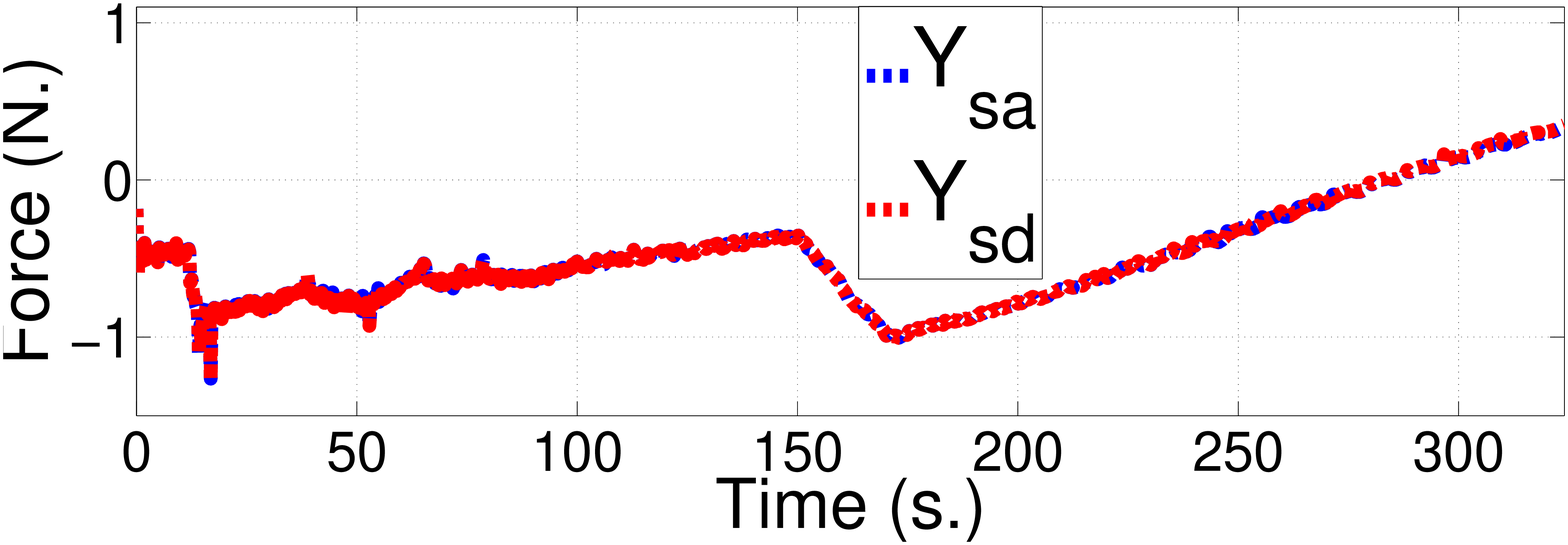}}
\subfigure[Output moment of thrust allocation in yaw]{\label{allpsi}\includegraphics[width=0.47\hsize]{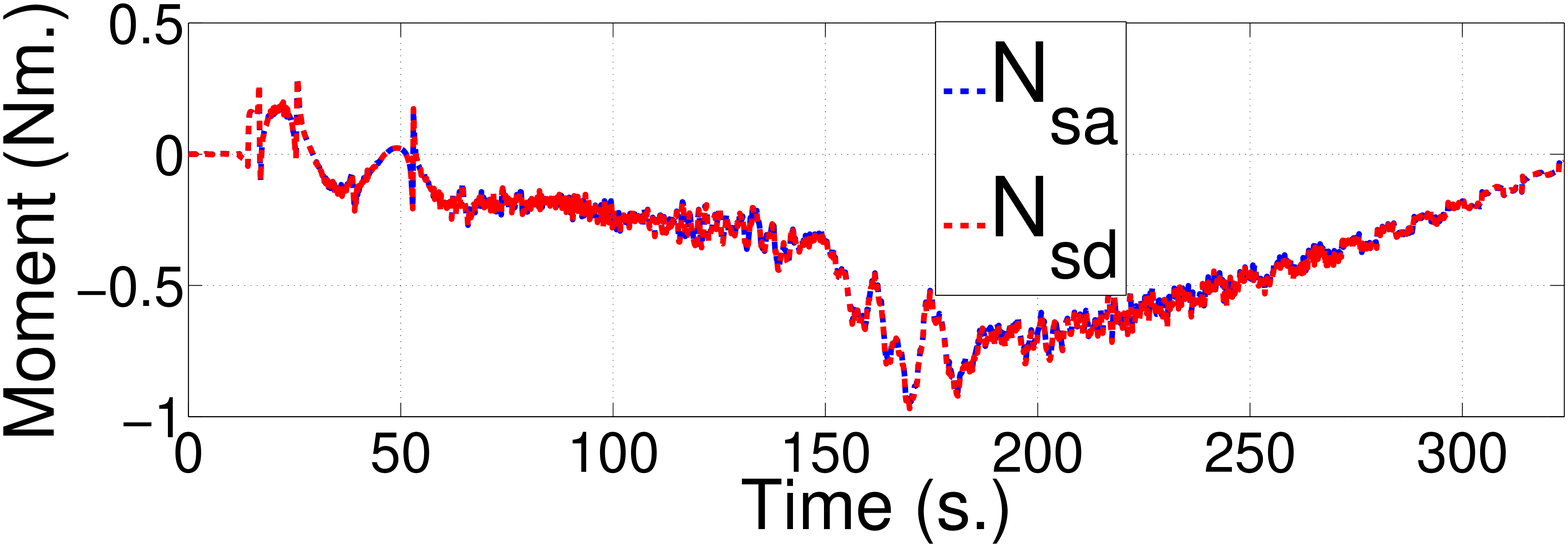}}
  \caption{Output of thrust allocation results.} \label{allresults}
\end{figure}

\begin{figure}[htb]
  \centering
  \includegraphics[width=90mm,height=35mm]{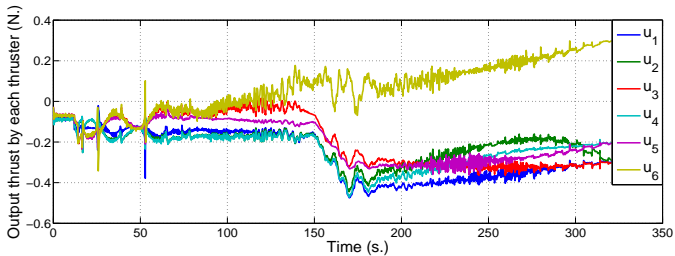}
  \caption{Output thrust by each thruster} \label{allu}
\end{figure}

\begin{figure}[htb]
  \centering
  \includegraphics[width=90mm,height=50mm]{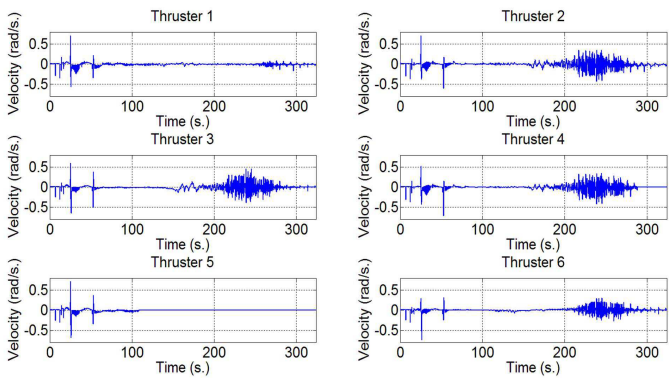}
  \caption{Angular velocity of each thruster} \label{anvel}
\end{figure}

\begin{figure}[htb]
  \centering
  \includegraphics[width=90mm,height=50mm]{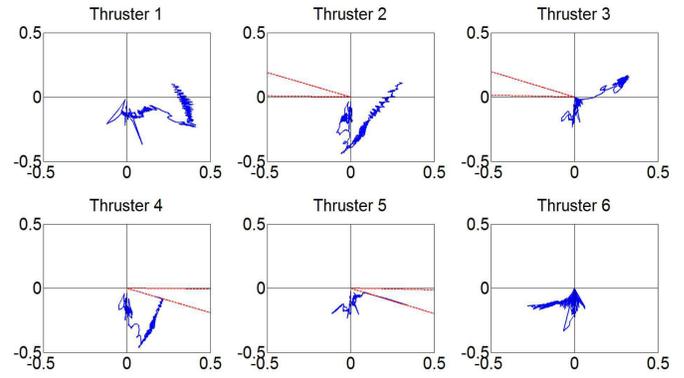}
  \caption{Azimuth angle and output trust tracking record} \label{thrtra}
\end{figure}

The configuration of the six thrusters are shown in Fig. \ref{AV2}. The encounter angles $\alpha_{e2}$,$\alpha_{e3}$,$\alpha_{e4}$ and $\alpha_{e5}$ are defined in Fig. \ref{encoan}. The specific values of the encounter angles are calculated as $\alpha_{e2}=190.9086^{\circ}, \alpha_{e3}=191.5165^{\circ}, \alpha_{e4}=10.8194^{\circ}, \alpha_{e5}=11.5165^{\circ}$.
Considering the forbidden zone of $20^{\circ}$, the working zone, in other words, the constraints for the azimuth angles are defined as $0^{\circ}\leq\alpha_1\leq360^{\circ}, 0^{\circ}\leq\alpha_2\leq180.8194^{\circ} \bigcup 200.8194^{\circ}\leq\alpha_2\leq360^{\circ}, 0^{\circ}\leq\alpha_3\leq181.5165^{\circ} \bigcup 201.5165^{\circ}\leq\alpha_3\leq360^{\circ}, 0^{\circ}\leq\alpha_4\leq0.8194^{\circ} \bigcup 20.8194^{\circ}\leq\alpha_4\leq360^{\circ}, 0^{\circ}\leq\alpha_5\leq1.5165^{\circ} \bigcup 21.5165^{\circ}\leq\alpha_5\leq360^{\circ}, 0^{\circ}\leq\alpha_6\leq360^{\circ}$ For the ease of calculation, we need to merge the separated subset of $\alpha_2$,$\alpha_3$,$\alpha_4$ and $\alpha_5$ into the following form.
\begin{align}
\notag &200.819^{\circ}\leq\alpha_2\leq540.819^{\circ},201.517^{\circ}\leq\alpha_3\leq541.517^{\circ}\\
&20.819^{\circ}\leq\alpha_4\leq360.819^{\circ}, 21.517^{\circ}\leq\alpha_5\leq361.517^{\circ}
\end{align}
Particularly, since $\alpha_1$ and $\alpha_6$ can achieve full round rotation, in simulation, we set no constraint of rotation angle for $\alpha_1$ and $\alpha_6$. The optimization weights $\mathcal{Q}$, $\mathcal{P}$ and $\mathcal{R}$ are selected as $\mathcal{Q}=0.2I_{6\times6}$, $\mathcal{P}=0.2I_{6\times6}$ and $\mathcal{R}=10I_{3\times3}$. The updating parameter $\Gamma_{\mathcal{Z}}$ in the dynamic solver (\ref{optiZ}) is tuned as $\Gamma_{\mathcal{Z}}=0.1I_{18\times18}$. The upper and lower bound of the variables in (\ref{optcon1}-\ref{optcon2}) are $\underline{u}=-0.7\text{ones}(6,1)$, $\overline{u}=0.7\text{ones}(6,1)$, $\underline{\Delta \alpha}=-\frac{\pi}{20}\text{ones}(6,1)/\Delta t_{\text{opt}}$, $\overline{\Delta \alpha}=\frac{\pi}{20}\text{ones}(6,1)/\Delta t_{\text{opt}}$. Where $\Delta t_{\text{opt}}=0.167$s is the sampling time interval between two loops. The constraint for the allocation error of the dynamic solver are $\underline{o}=-0.02\text{ones}(6,1)$ and $\overline{o}=0.02\text{ones}(6,1)$. The initial thrust that each thruster provides are $u_{0}=0.0308\text{ones}(6,1)$. The initial rotation angles are $\alpha_{0}=\left[\displaystyle\frac{\pi}{2}, \displaystyle\frac{5\pi}{2}, \displaystyle\frac{5\pi}{2}, \displaystyle\frac{\pi}{2}, \displaystyle\frac{\pi}{2}, \displaystyle\frac{\pi}{2}\right]^T$. To reduce the computation consumption, in practical implementation, a termination mechanism is introduced for each optimization loop. The maximum iteration number in each loop is $10^{5}$. If the variance of $J_c$ during the past 1000 successive iteration is smaller than $10^{-12}$, the convergence of current loop can be rationally identified and computation process is terminated. Fig. (\ref{allresults})-(\ref{thrtra}) show the simulation results of the dynamic allocator.

\subsection{Discussion}
Figs. \ref{olprate11} shows that the proposed control can handle the input delay under severely varying environmental circumstances. Good tracking performance is achieved under the hybrid feedforward and feedback control scheme in surge and sway. However, there is relatively large oscillation at the beginning of tracking in yaw direction, but the heading angle is able to converge to the desired trajectory gradually. The corresponding tracking errors are shown in Figs. \ref{olprate111}. It can be observed that all the tracking errors are successfully restricted within the predefined constraint $N_d$. The estimation of the peak of wind drag coefficient $\tilde{\Phi}$ is presented in Fig. \ref{wce}. As we can see, before the attack of the large wave-induced force, i.e. $t<150s$, the estimator is able to achieve accurate approximation. After the attack, the estimation values start to increase rapidly which help to trigger the compensation. The alarm is activated at 160.17s. After the alarm, the estimation values fall in overcompensation and the simulation curves are ignored in the plot become they do not have adequate actual meaning. The observation error for plane position in the sea state observer is necessarily small as shown in Fig. \ref{simdesterr}. However, larger observation error can be seen during 150s-160s due to the effect of the wave force. The large observer error vanishes after the involvement the NN compensator. This abrupt error variation can be employed as an auxiliary indicator to decide the alarm moment.\\
\indent In Figs. \ref{allresults}, the blue and red line represent the allocated generalized force and the command signal from the controller respectively. The results demonstrate that the dynamic allocator can provide satisfactory resulting force and moment to match the desired command signal. The produced thrust of each thruster is always within the limit of $\pm 0.7$N as shown in Fig. \ref{allu}. Combining Figs. \ref{anvel}-\ref{thrtra}, it is observed that constraints for the rotation angle and angular velocity are both not violated.
\section{Conclusion}
In this paper, DP control has been proposed for a marine vessel under uncertain environmental force variation due to adjoining FPSO. First, a novel sea state observer has been developed with adaptive wind force and moment estimator to alarm large wave-induced drift force. Then, the control system has been designed using SBLF and predictor-based method in combination with NN to handle the tracking error constraints, input delay as well as the unknown wave force. The stability of the proposed sea state observer and the controller has been shown through rigorous Lyapunov and Lyapunov-Krasovskii analysis respectively. Finally, dynamic thrust allocation has been sequently investigated for individual thrusters of the DP system employing locally convex reformulation and LVIPDNN method. Simulation study has been conducted to verify the effectiveness of the proposed control scheme and thrust allocation.
\ifCLASSOPTIONcaptionsoff
  \newpage
\fi

\scriptsize
\bibliographystyle{ieeetr}
\bibliography{ref}
\end{document}